\documentclass[10pt,twocolumn,letterpaper]{article}

\usepackage{wacv}                        

\usepackage{graphicx}    
\usepackage{amsmath}     
\usepackage{amssymb}     
\usepackage{booktabs}    
\usepackage{multirow}    
\usepackage[most]{tcolorbox} 

\newcommand{\red}[1]{{#1}}

\newtcolorbox{takeaway}{
  enhanced, unbreakable,
  colback=black!4, colframe=black, coltitle=black,
  boxrule=0.4pt, left=4pt, right=4pt, top=3pt, bottom=3pt,
  arc=2pt, boxsep=2pt,
  fonttitle=\bfseries\footnotesize, title={Key takeaway},
  attach title to upper=\ \,, fontupper=\footnotesize}

\definecolor{wacvblue}{rgb}{0.21,0.49,0.74}
\usepackage[pagebackref,breaklinks,colorlinks,allcolors=wacvblue]{hyperref}

\title{How Do Diffusion Classifiers Decide? A Bias-Centric Evaluation}

\author{
  Saba Fathi\textsuperscript{2} \quad
  Fardin Ayar\textsuperscript{2} \quad
  Maryam Abdolali\textsuperscript{1} \quad\\
  Ehsan Javanmardi\textsuperscript{3} \quad
  Manabu Tsukada\textsuperscript{3} \quad
  Mahdi Javanmardi\textsuperscript{2}\thanks{Corresponding author: \texttt{mjavan@aut.ac.ir}. \\
  \textsuperscript{1}\,\texttt{maryam.abdolali@kntu.ac.ir} \\
  \textsuperscript{2}\,\texttt{\{sabafathi, fardin.ayar, mjavan\}@aut.ac.ir} \\
  \textsuperscript{3}\,\texttt{\{ejavanmardi, mtsukada\}@g.ecc.u-tokyo.ac.jp}} \\
  \textsuperscript{1}K. N. Toosi University of Technology, Tehran, Iran \\
  \textsuperscript{2}Amirkabir University of Technology, Tehran, Iran \\
  \textsuperscript{3}The University of Tokyo, Tokyo, Japan
}

\begin{document}
\maketitle


\begin{abstract}
Diffusion models have recently been repurposed for zero-shot classification, giving rise to diffusion classifiers that identify the best-matching text prompt by minimizing the noise-prediction error. Despite their growing adoption, how these models make classification decisions remains poorly understood. We introduce \textbf{ASOB-Bench}, a bias evaluation for diffusion classifiers along three dimensions: \textbf{A}ttribute binding, \textbf{S}ize-\textbf{O}rder bias, and \textbf{B}ackground dependency. \red{These dimensions serve not as an exhaustive taxonomy but as targeted probes of \emph{how} the text-conditioned reconstruction-error score reaches a decision.} Such a perspective is well studied for discriminative vision-language models, yet remains overlooked for diffusion classifiers. Extending an existing framework with five new attribute categories on newly constructed datasets, we find diffusion classifiers are \emph{less} prone to attribute misbinding than an OpenCLIP baseline; on the established ComCo benchmark they are \emph{substantially more} susceptible to size-order shortcuts; and on ImageNet-B they suffer \emph{far larger} accuracy drops, revealing heavy reliance on background over foreground cues. Reconstruction-error heatmaps and U-Net cross-attention visualizations expose the mechanism behind each bias. Because diffusion classifiers share the same denoiser as text-to-image models, these single-pass diagnostics also point toward analogous failure modes in generation. Overall, diffusion classifiers exhibit a distinct bias profile from vision-language models, offering guidance for building more robust diffusion-based models.
\end{abstract}

\noindent Code and data to reproduce our experiments are publicly available at \url{https://github.com/sabafathi11/asob-bench}.


\begin{figure*}[t]
	\centering
	\includegraphics[width=\textwidth]{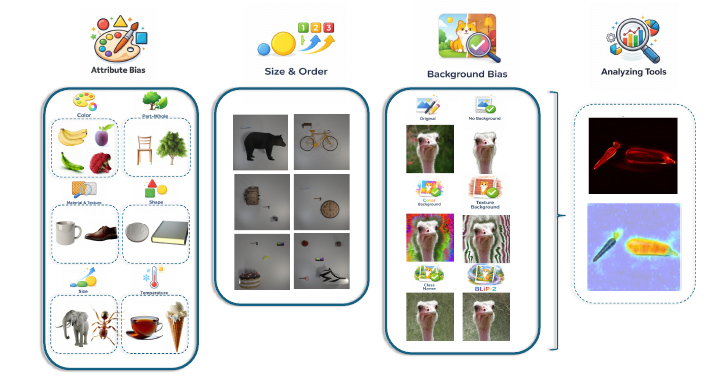}
	\caption{Overview of ASOB-Bench across three bias dimensions: attribute binding, size-and-order, and background dependency. Reconstruction-error heatmaps and cross-attention visualizations provide mechanistic insight into each bias.}
	\label{fig:overview}
\end{figure*}

\section{Introduction}
\label{sec:intro}

Diffusion models~\cite{Ho2020DenoisingDP} have emerged as the dominant paradigm in generative image modeling, synthesizing high-fidelity images from textual descriptions through learned iterative denoising. A recent insight is that these generative models can be repurposed for \emph{discriminative} tasks: by measuring how well a diffusion model can reconstruct a given image under each candidate text prompt, one obtains a zero-shot classifier with no task-specific training~\cite{Clark2023TexttoImageDM,Jeong2025DiffusionCU,Krojer2023AreDM,Li2023YourDM}. This generative-to-discriminative transfer has attracted growing interest, yet how these classifiers reach their decisions is still not well understood. One way to expose these mechanisms is through bias evaluation, which has proven effective in revealing systematic failure modes in discriminative vision-language models. For models such as CLIP~\cite{Radford2021LearningTV}, extensive work has documented failure modes ranging from attribute misbinding to spurious background reliance~\cite{Abdolali2026CompositionalitySurvey}. Diffusion classifiers have received far less attention despite their increasing adoption. Understanding their biases is therefore essential both for reliable deployment and, since diffusion classifiers share their generative mechanism with text-to-image models, for the broader diffusion landscape.

In this paper, we introduce \textbf{ASOB-Bench} and conduct a systematic evaluation along three complementary bias dimensions:
\begin{itemize}
	\item \textbf{Attribute binding bias}, the tendency to misassociate attributes with unintended objects in multi-object scenes;
	\item \textbf{Size-order bias}, preferential treatment of larger objects or earlier prompt tokens, conflating visual salience and token position with semantic correctness;
	\item \textbf{Background dependency bias}, over-reliance on background context rather than foreground object cues.
\end{itemize}

{ Each dimension isolates a distinct failure of the reconstruction-error score (\emph{cross-modal} grounding, \emph{spatial} error aggregation, and \emph{foreground--background} reliance) rather than forming an exhaustive taxonomy; axes such as prompt-length and plausibility~\cite{GoodCREPE2025}, or social bias~\cite{Krojer2023AreDM}, target prompt or dataset priors and lie outside our scope.}

Our contributions are threefold:
\begin{enumerate}
	\item We introduce \textbf{ASOB-Bench}, a multi-dimensional benchmark for evaluating biases in diffusion classifiers across attribute binding, size-order, and background dependency dimensions.
	\item { We provide the first systematic \emph{mechanistic} analysis of how the reconstruction-error score reaches a decision, using reconstruction-error heatmaps and cross-attention maps to trace each observed bias to a concrete cause.}
	\item { We report consistent empirical findings across these dimensions: diffusion classifiers exhibit \emph{less} attribute-binding bias than the baseline, yet are \emph{substantially more} susceptible to size-order shortcuts and background dependency.}
\end{enumerate}

{ More broadly, our study shows that headline accuracy reveals little about how a diffusion classifier actually decides, and that a bias-centric lens is needed to anticipate where these models will fail.}


\section{Related Work}
\label{sec:related_work}

\subsection{Diffusion Models and Diffusion Classifiers}

Diffusion models~\cite{Ho2020DenoisingDP}\red{\cite{song2020scorebased}} learn to generate data by reversing a forward process that progressively corrupts samples with noise. This framework achieves state-of-the-art results in image synthesis~\cite{Rombach2021HighResolutionIS,Saharia2022PhotorealisticTD}, video generation~\cite{Blattmann2023StableVD,Blattmann2023AlignYL}, and audio~\cite{Evans2024FastTL}, with Stable Diffusion~\cite{Rombach2021HighResolutionIS} as a widely adopted open-source text-to-image model.

A notable downstream application is zero-shot classification. Li et al.~\cite{Li2023YourDM} introduced the Diffusion Classifier, which selects the text prompt minimizing noise-prediction error, surpassing CLIP RN-50~\cite{Radford2021LearningTV} without task-specific training. Subsequent work has refined this paradigm: Clark et al.~\cite{Clark2023TexttoImageDM} propose a universal timestep-weighting scheme, Krojer et al.~\cite{Krojer2023AreDM} extend the idea to bidirectional retrieval via Diffusion-ITM, and a recent study~\cite{Jeong2025DiffusionCU} highlights sensitivity to dataset domain and timestep selection. \red{More recent efforts optimize the matching noise used for scoring~\cite{NoiseMatters2025} or parameterize few-shot learners by diffusion timesteps~\cite{Yue2024FewShotDiffusionTimesteps,He2023DiscffusionDD}.} While these works establish diffusion classifiers as competitive discriminative models, they focus on accuracy rather than systematic failure modes, leaving bias largely unaddressed.

\subsection{Bias and Compositionality in Vision Models}

Compositionality, the ability to correctly combine objects, attributes, and spatial relations, remains a core challenge for both generative and discriminative vision models. In the generative setting, diffusion models struggle with out-of-distribution attribute combinations~\cite{Okawa2023CompositionalAE}, and benchmarks such as GenEval~\cite{Ghosh2023GenEvalAO} and T2I-CompBench++~\cite{Huang2023T2ICompBenchAE} show persistent failures in spatial reasoning, attribute binding, and numeracy. On the discriminative side, CLIP~\cite{Radford2021LearningTV} frequently misbinds attributes and mishandles spatial relations due to its contrastive objective favoring global over fine-grained alignment~\cite{Abdolali2026CompositionalitySurvey,Ma2022CC,Wang2025Spatial457AD,Yuksekgonul2022WhenAW}. Recent evidence suggests diffusion classifiers can outperform CLIP on certain compositional tasks~\cite{Jeong2025DiffusionCU}.

Compositionality benchmarks establish \emph{what} models fail at; bias evaluation probes \emph{why} by tracing failures to systematic, data-driven shortcuts. Prior work has documented several relevant categories: \textit{Attribute binding bias}, Tang et al.~\cite{Tang2022WhenAL} formalize Concept Association Bias (CAB), showing models default to prototypical attribute--object pairings in multi-object scenes. \textit{Size and spatial bias}, Abbasi et al.~\cite{Abbasi2025CLIPUT} demonstrate that vision-language encoders preferentially attend to larger objects, while generative models struggle with positional relationships~\cite{Wu2024ConceptMixAC,Han2025SpatialTO}. \textit{Background dependency}, Malik et al.~\cite{Malik2024ObjectComposeER} show that altering background context substantially degrades classification accuracy.

Despite this growing body of work, existing bias analyses focus primarily on either discriminative vision-language models or generative image synthesis systems. Diffusion classifiers lie at the intersection of these paradigms: they use generative models to perform discriminative inference. However, their bias characteristics and decision mechanisms remain largely unexplored. \red{The closest prior effort, GDBench~\cite{Krojer2023AreDM}, probes social and attribute bias of diffusion models through image--text matching; in contrast, we target the \emph{decision mechanism} of the reconstruction-error score itself and trace, at the pixel and timestep level, \emph{why} each bias arises. In this work, we provide the first systematic mechanistic investigation of how this scoring mechanism produces such behaviors.}


\section{Preliminaries}
\label{sec:preliminaries}

\subsection{Model Configuration}

\paragraph{Diffusion classifiers.}
For an image--text pair $(x, c)$, the image $x$ is encoded into latent $z$ via a pretrained autoencoder, and the model is trained to predict the noise $\epsilon$ added to corrupted latent $z_t$ at timestep $t$:
\begin{equation}
	\label{eq:diffusion_loss}
	\mathcal{L}(z, c) = \mathbb{E}_{t,\epsilon}\!\left[w_t\,\bigl\|\epsilon - \epsilon_\Theta(z_t, t, c)\bigr\|^2\right],
\end{equation}
where $w_t$ are timestep weights and $\epsilon_\Theta$ is the denoising network conditioned on prompt $c$. This objective relates to the ELBO of $p(z \mid y)$, allowing diffusion models to be repurposed for zero-shot classification~\cite{Clark2023TexttoImageDM,Krojer2023AreDM,Li2023YourDM}: given candidate class labels $\{y_k\}$, the predicted class is
\begin{equation}
	\label{eq:diffusion_classifier}
	\tilde{y} = \arg\max_{y_k}\, p(y = y_k \mid z) = \arg\max_{y_k}\, \log p(z \mid y = y_k),
\end{equation}
approximated by selecting the prompt that minimizes the noise-prediction error in Eq.~\eqref{eq:diffusion_loss}.

We build our diffusion classifier on Stable Diffusion~2~\cite{Rombach2021HighResolutionIS}, following Li et al.~\cite{Li2023YourDM}, with uniform sampling over 50 timesteps and no pruning, as our goal is to systematically analyze bias rather than optimize for peak accuracy or computational efficiency. For attribute tasks we restrict sampling to timesteps 800--999~\cite{Jeong2025DiffusionCU}. As baseline, we use OpenCLIP ViT-H/14, which shares the same text encoder as Stable Diffusion~2, isolating the effect of the classification mechanism (contrastive matching vs.\ reconstruction-error scoring) rather than text encoding.

\subsection{Evaluation Framework}

\textbf{(1) Attribute binding bias.} A model must correctly associate an attribute with the intended object when multiple objects coexist in a scene. Tang et al.~\cite{Tang2022WhenAL} formalize this challenge as Concept Association Bias (CAB): models that process inputs as unordered bags of concepts tend to fill in missing attribute--object pairings crossmodally, defaulting to prototypical associations learned during training. For instance, given an image of a strawberry and a banana, a biased model may choose the prompt ``a yellow strawberry'' over the correct ``a red strawberry,'' since the banana's color already explains the yellow in the scene. Following~\cite{Tang2022WhenAL}, we quantify this tendency with:
\begin{equation}
	\label{eq:cab}
	\text{CAB} = \frac{\text{two-object*} - \text{two-object} + 1}{2}
\end{equation}
where \textit{two-object} denotes accuracy when the model correctly binds the attribute to the queried object, and \textit{two-object*} denotes accuracy when it instead predicts the \emph{other} object's attribute (i.e., systematic misbinding). The metric maps the gap between these two quantities to $[0,1]$: high \textit{two-object*} paired with low \textit{two-object} produces values near~1, indicating severe bias. We report four complementary metrics: \textit{Single} (attribute recognition in isolation), \textit{Two-object} (correct binding with a distractor present), \textit{Two-object*} (systematic misbinding rate), and \textit{CAB} (Eq.~\ref{eq:cab}). { We want to emphasize that CAB is a diagnostic from Tang et al.~\cite{Tang2022WhenAL}, not a contribution of ours. Since it measures only the \emph{direction} of misbinding, we always read it jointly with the single- and two-object accuracies (Sec.~\ref{sec:experiments}).}

To prevent word-order effects from contaminating the attribute-binding signal~\cite{Abbasi2025CLIPUT}, we adopt simplified prompts (e.g., ``a red strawberry and another object''). A systematic comparison of 12 candidate prompt formulations is provided in Appendix~\ref{appendix:prompts}.

\begin{table*}[!t]
	\centering
	\caption{Attribute binding evaluation (\textit{Single}, \textit{Two objects}, \textit{Two objects*} are accuracies; \textit{CAB} is the misbinding diagnostic of Eq.~\ref{eq:cab}, where higher means more bias). Color values are proportions (0--1); other categories are percentages. \red{The ``Size Attribute'' row probes binding of size adjectives (e.g., \emph{big}/\emph{small}) and is distinct from the size-order shortcut study of Sec.~\ref{sec:experiments} (Table~\ref{tab:size_eval}).}}
	\label{tab:attribute_results}
	\small
	\begin{tabular}{llcccc}
		\toprule
		Category & Model & Single & Two objects & Two objects* & CAB \\
		\midrule
		\multirow{2}{*}{Color (natural)}
		& OpenCLIP ViT-H/14   & 0.958 & 0.112 & 0.867 & 0.8775 \\
		& Diffusion classifier & 0.776 & 0.212 & 0.64  & 0.714  \\
		\midrule
		\multirow{2}{*}{Color (unnatural)}
		& OpenCLIP ViT-H/14   & 1.0 & 0.619 & 0.361 & 0.371 \\
		& Diffusion classifier & 1.0 & 0.615 & 0.339 & 0.362 \\
		\midrule
		\multirow{2}{*}{Part-Whole}
		& OpenCLIP ViT-H/14   & 35.64 & 9.5  & 60.2 & 0.7535 \\
		& Diffusion classifier & 35.64 & 11.2 & 16   & 0.524  \\
		\midrule
		\multirow{2}{*}{Material/Texture}
		& OpenCLIP ViT-H/14   & 64.36 & 25.9 & 44.5 & 0.593  \\
		& Diffusion classifier & 36.63 & 7.9  & 15.6 & 0.5385 \\
		\midrule
		\multirow{2}{*}{Shape}
		& OpenCLIP ViT-H/14   & 98.02 & 38.18 & 61.9 & 0.6186 \\
		& Diffusion classifier & 81.19 & 55.4  & 44.6 & 0.446  \\
		\midrule
		\multirow{2}{*}{Size Attribute}
		& OpenCLIP ViT-H/14   & 82 & 33.8 & 66.2 & 0.662 \\
		& Diffusion classifier & 62 & 53.3 & 46.7 & 0.467 \\
		\midrule
		\multirow{2}{*}{Temperature}
		& OpenCLIP ViT-H/14   & 77 & 25.8 & 74.2 & 0.742 \\
		& Diffusion classifier & 81 & 48.7 & 51.3 & 0.513 \\
		\bottomrule
	\end{tabular}
\end{table*}

\textbf{(2) Size relationship and word-order bias.} Beyond attribute binding, models may exploit shortcut cues tied to visual salience or token position. Abbasi et al.~\cite{Abbasi2025CLIPUT} identify two complementary encoder-level biases: the image encoder tends to favor objects that occupy larger image regions, since more patches receive more attention, while the text encoder overweights objects mentioned earlier in the prompt, a pattern reinforced by training corpora where larger objects are typically described first. We adopt the ComCo evaluation protocol from~\cite{Abbasi2025CLIPUT}, which provides controlled multi-object images (2--5 objects) with one noticeably larger object, and defines two diagnostic scenarios. { Throughout, the \emph{positive} prompt is the caption that correctly matches the image, while a \emph{negative} prompt is a minimally altered caption that changes exactly one object; a correct classifier should assign higher reconstruction error to the negative prompt.} In \textit{Scenario~1} (bias-aligned), the correct prompt names the largest object first and the negative prompt changes that large object; because the changed region is also the largest, it is the most visually prominent and the model can correctly identify and reject the negative prompt. Size cues therefore favor the correct answer, so high accuracy may still reflect shortcuts rather than genuine understanding. In \textit{Scenario~2} (bias-conflicting), the correct prompt places the largest object last and the negative prompt changes a smaller object; the large unchanged object dominates the scene by area, overshadowing the semantically relevant change and making the negative prompt score competitively, causing accuracy to collapse. Any size or primacy shortcut thus actively pushes toward the wrong answer. The accuracy gap between the two scenarios directly quantifies shortcut vulnerability.

\textbf{(3) Background dependency bias.} \mbox{ImageNet-B}~\cite{Malik2024ObjectComposeER} provides images under five conditions: \emph{original} (the unaltered image), \emph{class-name} and \emph{BLIP} (the background regenerated from the object's class name or from a BLIP caption), and \emph{color} and \emph{texture} (the background altered to conflict in color or texture). We add a sixth \emph{no-background} condition, the segmented foreground alone, as an upper-bound foreground-only reference.

\begin{figure*}[t]
	\centering
	\includegraphics[width=0.80\textwidth]{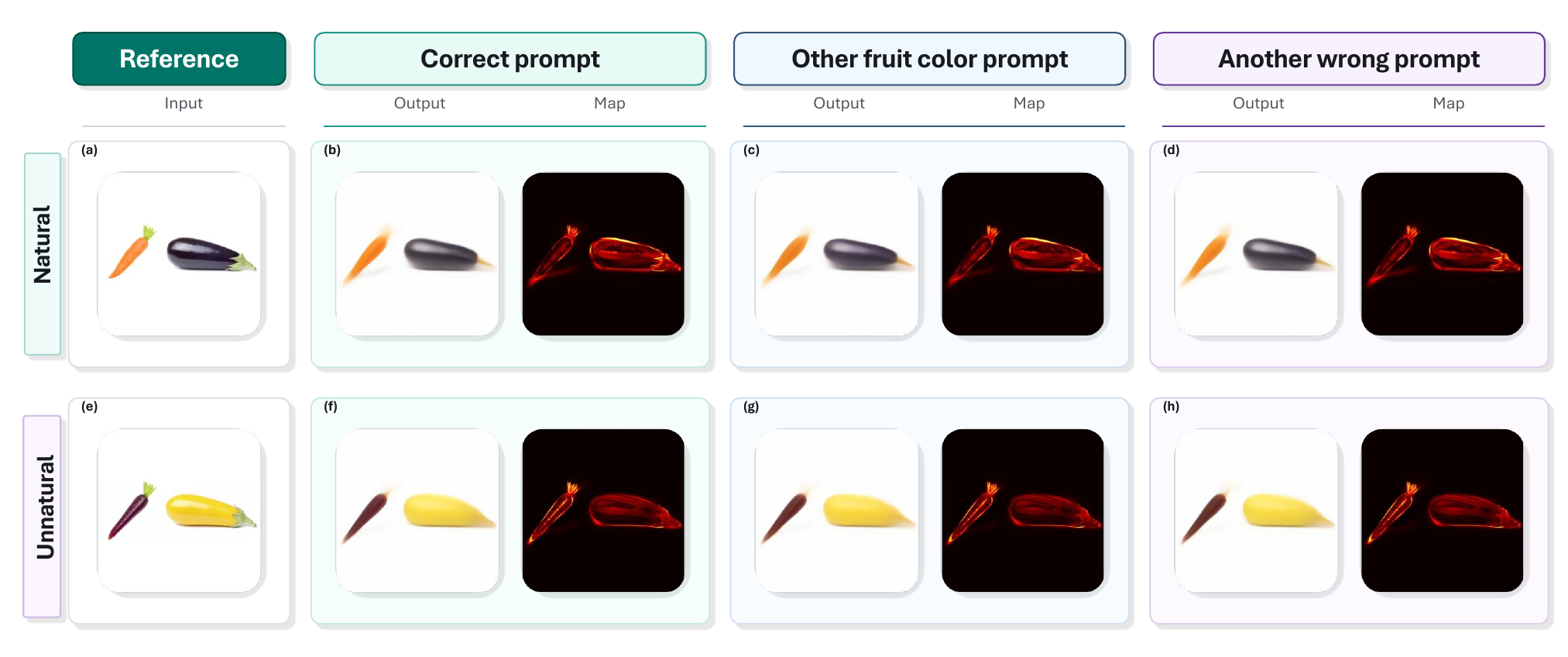}
	\caption{One-step reconstruction error heatmaps for natural (top) and unnatural (bottom) color pairs. On natural images, error concentrates on the distractor eggplant. On unnatural images, error correctly shifts to the queried carrot, confirming misbinding stems from learned color--object co-occurrences.}
	\label{fig:recon_error_color}
\end{figure*}

\subsection{Datasets}

\textbf{Attribute binding.} We evaluate color binding using two datasets: a natural-color fruit dataset and a synthetic dataset generated with DALL-E~3~\cite{Betker2023DallE3} containing unnatural color combinations (e.g., orange strawberries, purple bananas). These datasets allow us to distinguish between biases arising from learned co-occurrence statistics and those arising from structural model behavior. { The natural-color fruit set follows the color-binding setup of prior work~\cite{Tang2022WhenAL}, whereas the unnatural-color DALL-E~3 set is introduced here (Fig.~\ref{fig:overview}). Beyond color, we further introduce five additional attribute categories (part--whole, material/texture, shape, size, and temperature), each as a two-object scene pairing a queried object with a distractor (additional samples and definitions in Appendix~\ref{appendix:datasets}).}

\textbf{Size and word-order.} We use the ComCo dataset~\cite{Abbasi2025CLIPUT}, which contains controlled multi-object scenes with varying object sizes and spatial arrangements {(Fig.~\ref{fig:overview})}. { We adopt its scenes unchanged, but since scoring a single image requires evaluating the noise-prediction error of every candidate prompt over 50 sampled timesteps (Eq.~\ref{eq:diffusion_loss}), we evaluate on a subsample; the exact counts are listed in Appendix~\ref{appendix:datasets}.}

\textbf{Background dependency.} We use ImageNet-B~\cite{Malik2024ObjectComposeER} {(Fig.~\ref{fig:overview})}, { adopted unchanged from prior work,} and { additionally introduce a no-background, foreground-only variant} by segmenting objects with BiRefNet~\cite{Zheng2024BiRefNet}. { We evaluate all six conditions on a focused subset of 20 high-quality classes, enumerated in Appendix~\ref{appendix:datasets}.}


\section{Experiments}
\label{sec:experiments}

\subsection{Attribute Bias Evaluation}

\subsubsection{Color Attribute Results}

Table~\ref{tab:attribute_results} (color, natural) shows that the diffusion classifier lags on single-object recognition (0.776 vs.\ 0.958) but is markedly less prone to systematic misbinding: it achieves higher two-object accuracy (0.212 vs.\ 0.112), lower two-object* (0.64 vs.\ 0.867), and a lower CAB (0.714 vs.\ 0.8775). The diffusion classifier makes more generic recognition errors, but when forced to choose between two candidates it is less likely to default to the distractor's stereotypical color.

On unnatural fruit colors (Table~\ref{tab:attribute_results}, color, unnatural), the CAB effect largely vanishes: both models achieve perfect single-object accuracy and correct-binding accuracy exceeds misbinding for both. This reversal is consistent with the CAB analysis of Tang et al.~\cite{Tang2022WhenAL}: breaking prototypical associations removes the shortcut, causing the systematic two-object vs.\ two-object* gap to collapse.

\subsubsection{Additional Attribute Evaluations}

According to Table~\ref{tab:attribute_results}, across all five additional categories, the diffusion classifier consistently achieves a lower CAB than the baseline. Both models exhibit attribute binding bias, yet the diffusion classifier is less biased in every category. { Crucially, this lower CAB is paired with \emph{higher} two-object accuracy in every category, confirming genuinely fewer misbindings rather than weaker recognition.}

\subsubsection{Visualization Analysis for Attribute Bias}

\begin{figure*}[tb]
	\centering
	\begin{minipage}[t]{0.48\textwidth}
		\centering
		\includegraphics[width=\linewidth]{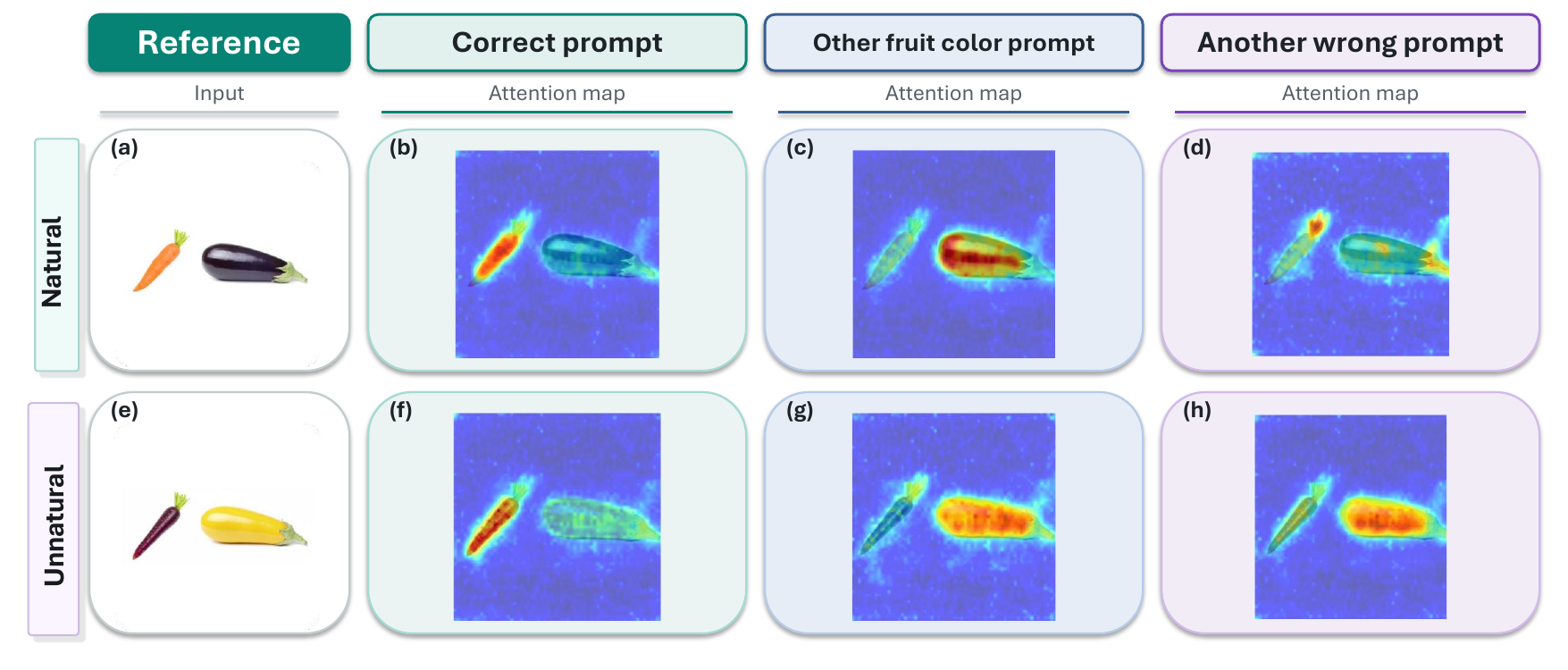}
		\caption{Cross-attention heatmaps for ``carrot'' under natural (top) and unnatural (bottom) conditions. Attention is routed by color cue regardless of prototypical associations, confirming this is structural to the cross-attention mechanism.}
		\label{fig:attention_color}
	\end{minipage}\hfill
	\begin{minipage}[t]{0.48\textwidth}
		\centering
		\includegraphics[width=\linewidth]{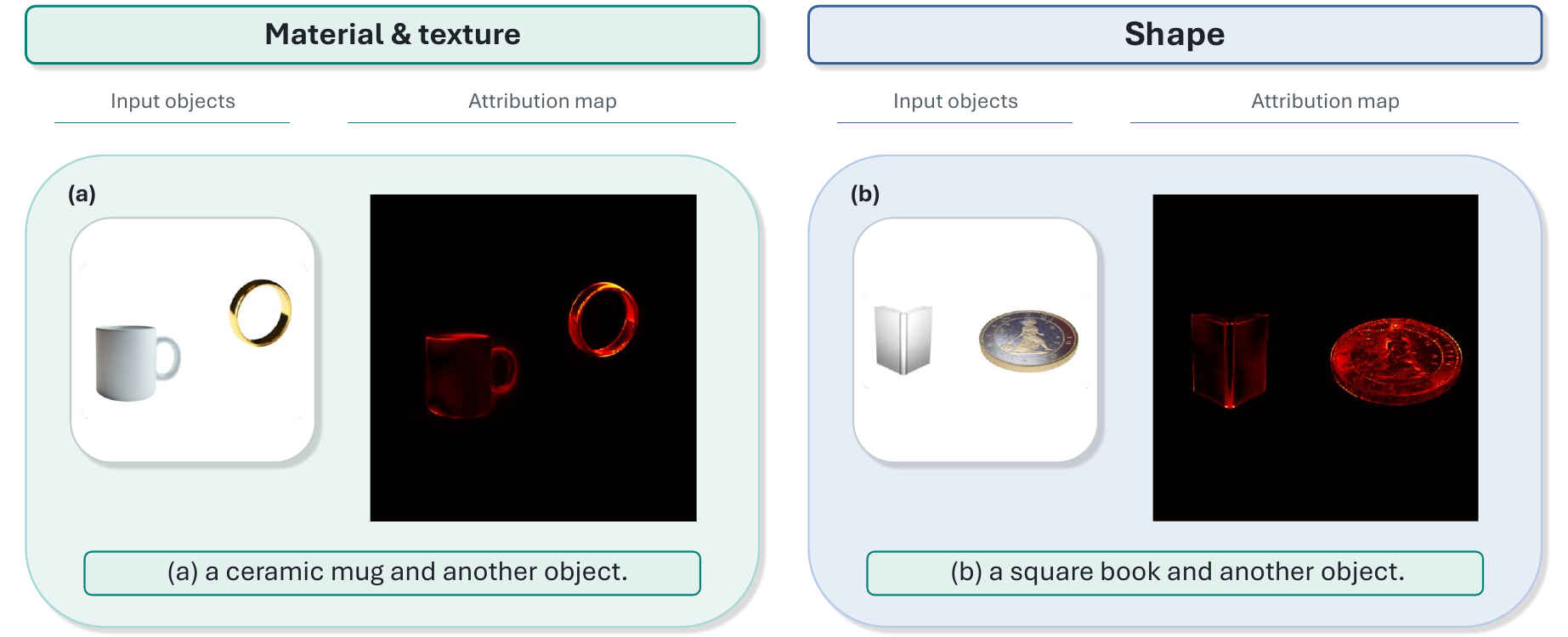}
		\caption{Error heatmaps for non-color attributes: (a)~material/texture: error on the ring (distractor); (b)~shape: error on the coin (distractor).}
		\label{fig:other_attr_heatmap}
	\end{minipage}
\end{figure*}

\begin{figure*}[tb]
	\centering
	\includegraphics[width=0.80\textwidth]{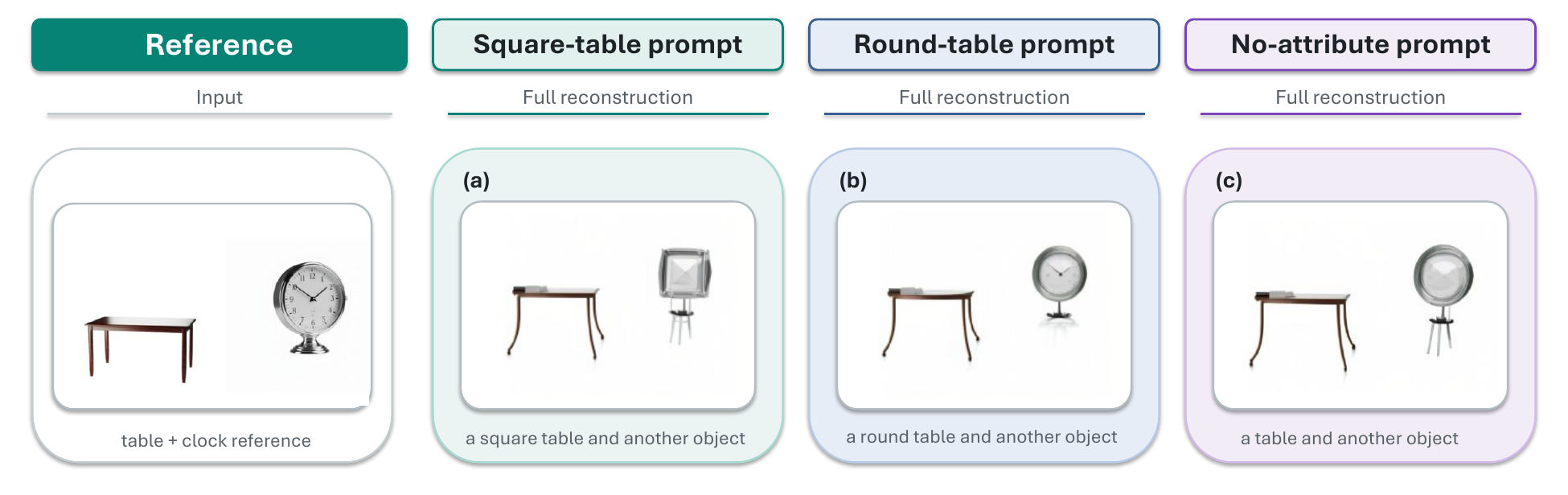}
	\caption{\red{Full reconstruction for a table-and-clock scene. \emph{Reference}: the input table and clock. (a)~Square-table prompt: square forms leak onto the round clock. (b)~Round-table prompt: the clock is already round, so the attribute is satisfied without visible change. (c)~No-attribute prompt: no leakage appears.}}
	\label{fig:full_recon}
\end{figure*}

\paragraph{Reconstruction-error heatmaps reveal misattributed binding.}

We perform a single denoising step under each candidate prompt, decode the resulting latent to pixel space, and compute a per-pixel difference map against the original image. This error heatmap indicates where the model perceives a prompt--image mismatch, i.e., the regions it would modify to better align the two. Figure~\ref{fig:recon_error_color} presents these heatmaps for three prompt types: the \emph{correct prompt} (e.g., ``an orange carrot and another object''), an \emph{other-fruit-color prompt} naming the distractor's color (e.g., ``a purple carrot and another object''), and an \emph{absent-color prompt} specifying a color not present in the scene (e.g., ``a green carrot and another object'').

On natural images (Figure~\ref{fig:recon_error_color}, top row), reconstruction error is consistently concentrated on the eggplant, the object \emph{not} queried by the prompt, rather than on the carrot. This pattern persists across all three prompt types, exposing a characteristic CAB failure mode: because the carrot's prototypical orange color is already ``explained'' by its object token, the dominant reconstruction discrepancy shifts to the distractor. This provides a mechanistic account of the high two-object* scores in Table~\ref{tab:attribute_results} (color, natural): prompts that incorrectly bind the attribute to the distractor can yield lower total reconstruction error than prompts requiring the model to override the prototypical association.

On unnatural images (Figure~\ref{fig:recon_error_color}, bottom row), where colors have been swapped, the error shifts to the carrot, the queried object, across all prompt types. With prototypical associations broken, the model can no longer rely on the object-token-to-color shortcut, and reconstruction error correctly reflects the intended binding. This is consistent with the collapse of the two-object vs.\ two-object* gap in Table~\ref{tab:attribute_results} (color, unnatural).

\paragraph{Cross-attention heatmaps expose color-driven routing.}
We visualize U-Net cross-attention maps aggregated across all layers for the token ``carrot,'' following DAAM~\cite{Tang2023DAAM}. Figure~\ref{fig:attention_color} reveals that the model routes ``carrot'' to whichever region best matches the \emph{color} in the prompt: under ``a purple carrot,'' attention shifts to the eggplant; under ``an orange carrot,'' it correctly localizes the carrot. This persists for both natural and unnatural images, confirming it is an intrinsic property of the cross-attention mechanism rather than a co-occurrence artifact.

\begin{figure*}[tb]
	\centering
	\begin{minipage}[t]{0.48\textwidth}
		\centering
		\includegraphics[width=\linewidth]{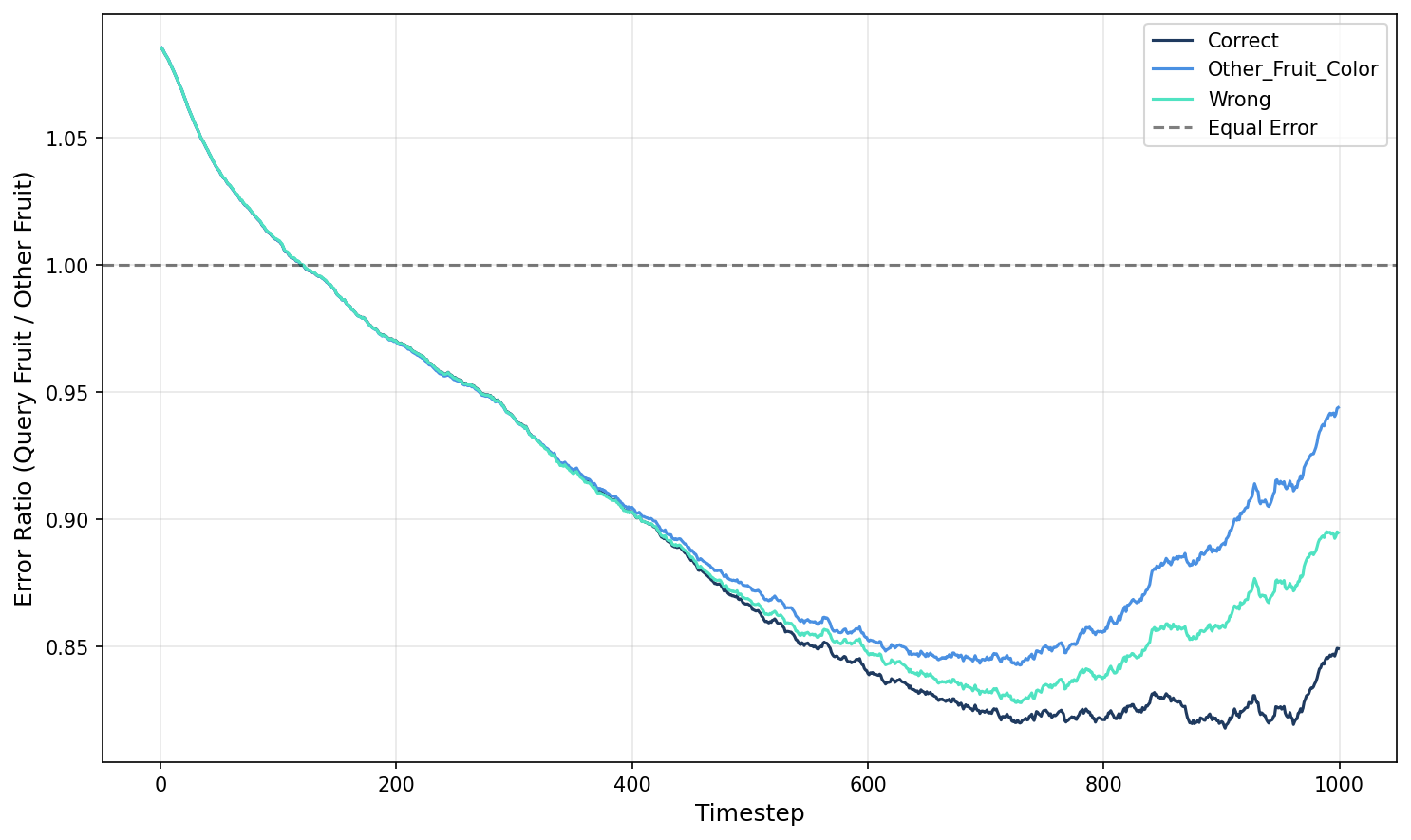}
		\caption{Error ratio (query/distractor) on natural images. In the attribute-binding regime ($t \geq 600$), all prompt types fall below~1, with bias deepening at higher timesteps.}
		\label{fig:error_ratio_natural}
	\end{minipage}\hfill
	\begin{minipage}[t]{0.48\textwidth}
		\centering
		\includegraphics[width=\linewidth]{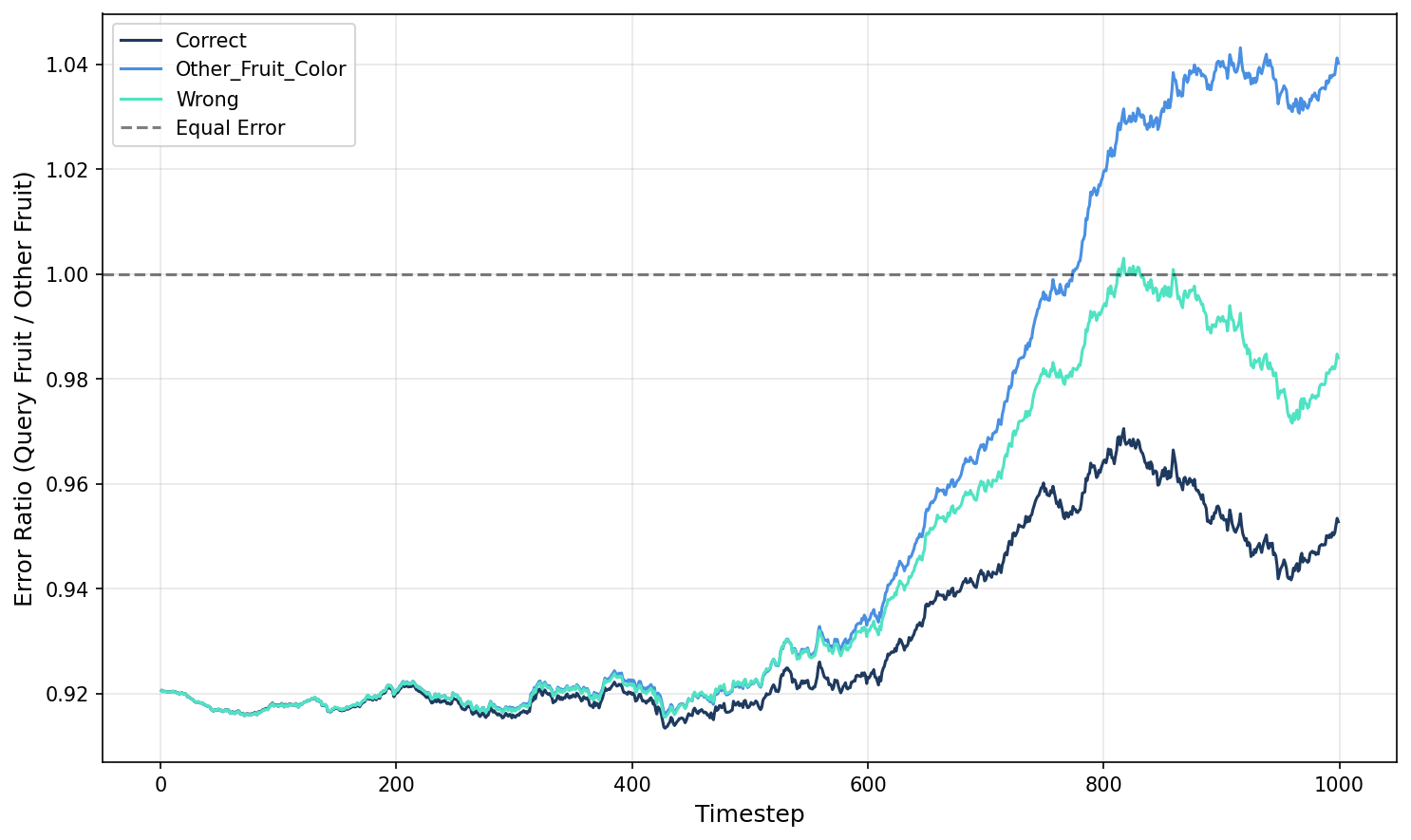}
		\caption{Error ratio (query/distractor) on unnatural images. Above $t \approx 600$ the ratio exceeds~1, confirming error correctly shifts to the query object once prototypical associations are broken.}
		\label{fig:error_ratio_unnatural}
	\end{minipage}
\end{figure*}

\paragraph{Generalization to other attribute categories.}

The same error-on-distractor pattern observed for color binding also occurs in other attribute categories. Figure~\ref{fig:other_attr_heatmap} illustrates this for shape and material/texture.

\paragraph{Full reconstruction confirms attribute leakage to the distractor.}

\red{The single-step heatmaps show only that reconstruction error is \emph{large on} the distractor, which does not by itself prove the attribute is bound there. A full reconstruction removes this ambiguity: by running the scheduler through all denoising timesteps, we let the model render its prediction and observe whether the queried attribute physically migrates onto the distractor.} Figure~\ref{fig:full_recon} shows a table-and-clock scene: prompted with ``a square table and another object'' (Fig.~\ref{fig:full_recon}(a)), squared forms appear on the clock while the table remains relatively unchanged; under ``a round table and another object'' (Fig.~\ref{fig:full_recon}(b)), the clock is already round, so the attribute is satisfied without visible change. Repeating with the attribute omitted eliminates the leakage entirely (Fig.~\ref{fig:full_recon}(c)), confirming the distortions arise from biased attribute binding rather than generic reconstruction artifacts.

\paragraph{Timestep dynamics of error and attention bias.}

To characterize how bias evolves across diffusion, we compute the per-timestep ratio of reconstruction error (and cross-attention) in the query-object region to the distractor region. A ratio below 1 means the model concentrates more error on the distractor than the queried object, the hallmark of misbinding; above 1, error correctly tracks the queried object. On natural images (Figure~\ref{fig:error_ratio_natural}), the ratio drops below~1 in the attribute-binding regime ($t \geq 600$) for all prompt types, with misbinding deepening toward $t \approx 800$, precisely the range Jeong et al.~\cite{Jeong2025DiffusionCU} recommend for improved attribute accuracy, so the timestep selection that improves accuracy also amplifies misbinding. On unnatural images (Figure~\ref{fig:error_ratio_unnatural}), the pattern reverses: above $t \approx 600$ the ratio climbs above~1, confirming that breaking prototypical associations causes reconstruction error to correctly track the queried object. The attention-ratio analysis (Appendix~\ref{appendix:att_ratios}) shows near-identical patterns for natural and unnatural images. This reveals a key dissociation: reconstruction-error bias is tied to learned co-occurrences and reverses when they are broken, while attention-routing bias is structural and cannot be disrupted simply by using unnatural colors.

\begin{takeaway}
Diffusion classifiers misbind attributes \emph{less} than the OpenCLIP baseline across all seven categories (lower CAB with \emph{higher} two-object accuracy), and reconstruction-error/attention heatmaps trace this to learned color--object co-occurrences, an error component that reverses once prototypical associations are broken.
\end{takeaway}

\subsection{Size and Order Evaluation}

Table~\ref{tab:size_eval} presents results averaged across all object counts and spatial positions. Both models perform strongly in Scenario~1 (baseline: 0.9967; diffusion: 0.9528), confirming that aligned salience and word-position cues do not compromise accuracy. Scenario~2 reveals sharp degradation: the diffusion classifier falls to 0.5853, substantially below the baseline at 0.7676.

\begin{table}[!tb]
	\centering
	\caption{Size and order evaluation (accuracy, averaged across all object counts (2--5) and spatial positions; higher is better). \red{The diffusion classifier's larger gap indicates greater susceptibility.}}
	\label{tab:size_eval}
	\small
	\begin{tabular}{lccc}
		\toprule
		Model & Scenario 1 & Scenario 2 & \red{Gap} \\
		\midrule
		Diffusion classifier & 0.9528 & 0.5853 & \red{0.3675} \\
		OpenCLIP ViT-H/14   & 0.9967 & 0.7676 & \red{0.2291} \\
		\bottomrule
	\end{tabular}
\end{table}

\paragraph{Reconstruction-error heatmaps reveal area-weighted shortcut.}

\begin{figure}[tb]
	\centering
	\includegraphics[width=\linewidth]{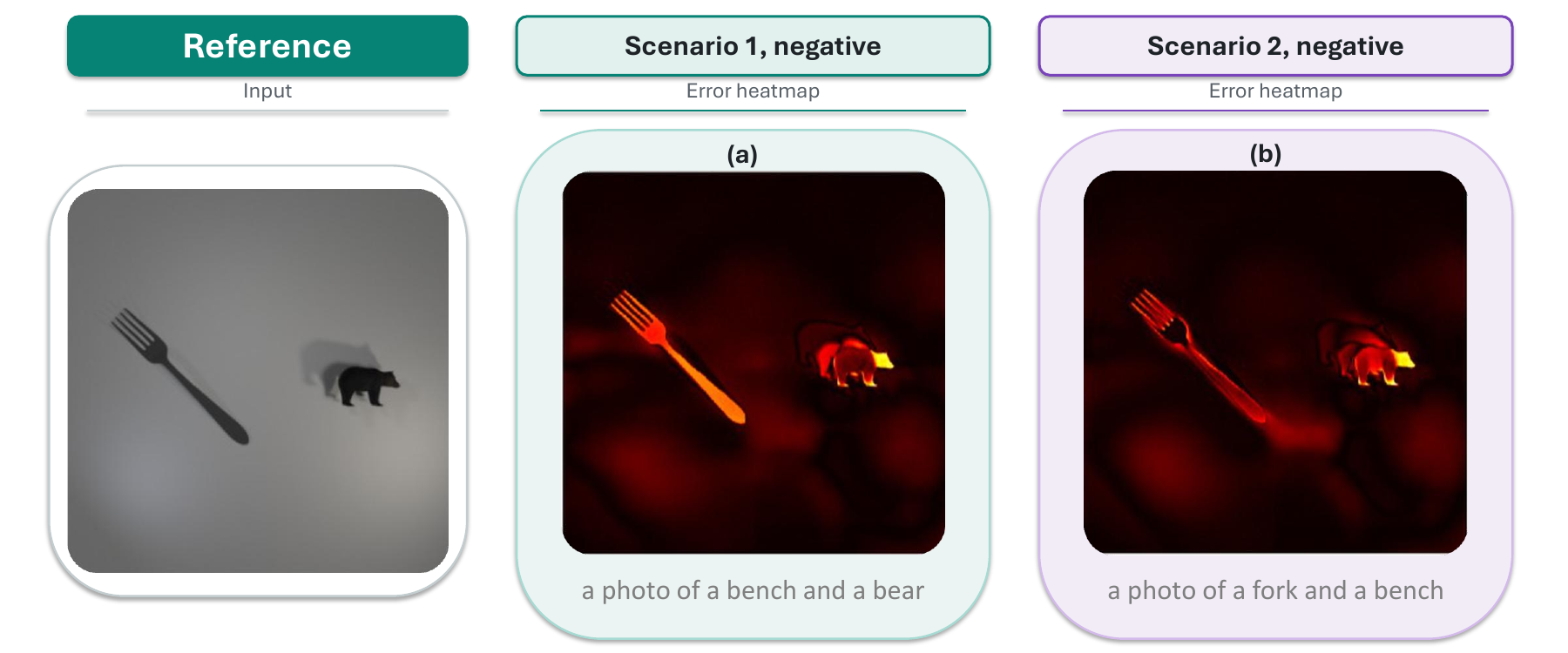}
	\caption{Error heatmaps for negative prompts of Scenario~1 (a) and Scenario~2 (b). Error correctly concentrates on the changed object, yet the fork's large spatial footprint generates sufficient residual error to make the negative prompt competitive.}
	\label{fig:size_neg}
\end{figure}

The mechanism is transparent: the classifier selects the prompt minimizing \emph{total} pixel-aggregated reconstruction error, so any larger region contributes more to the global score regardless of semantic relevance. Figure~\ref{fig:size_neg} shows that while error correctly concentrates on the changed object, the fork's large spatial footprint accumulates sufficient residual error to allow the negative prompt to score competitively. A timestep-resolved analysis (Appendix~\ref{appendix:size_ratios}) confirms the bias is driven primarily by object size rather than word order.

\begin{takeaway}
On size-order, the diffusion classifier is \emph{far more} susceptible to the shortcut than the OpenCLIP baseline (gap of 0.37 vs.\ 0.23). Because prompt selection minimizes \emph{total} pixel-aggregated error, large objects dominate the score regardless of semantic relevance, an area-weighted shortcut intrinsic to the scoring rule.
\end{takeaway}

\begin{figure*}[!t]
	\centering
	\includegraphics[width=0.68\textwidth]{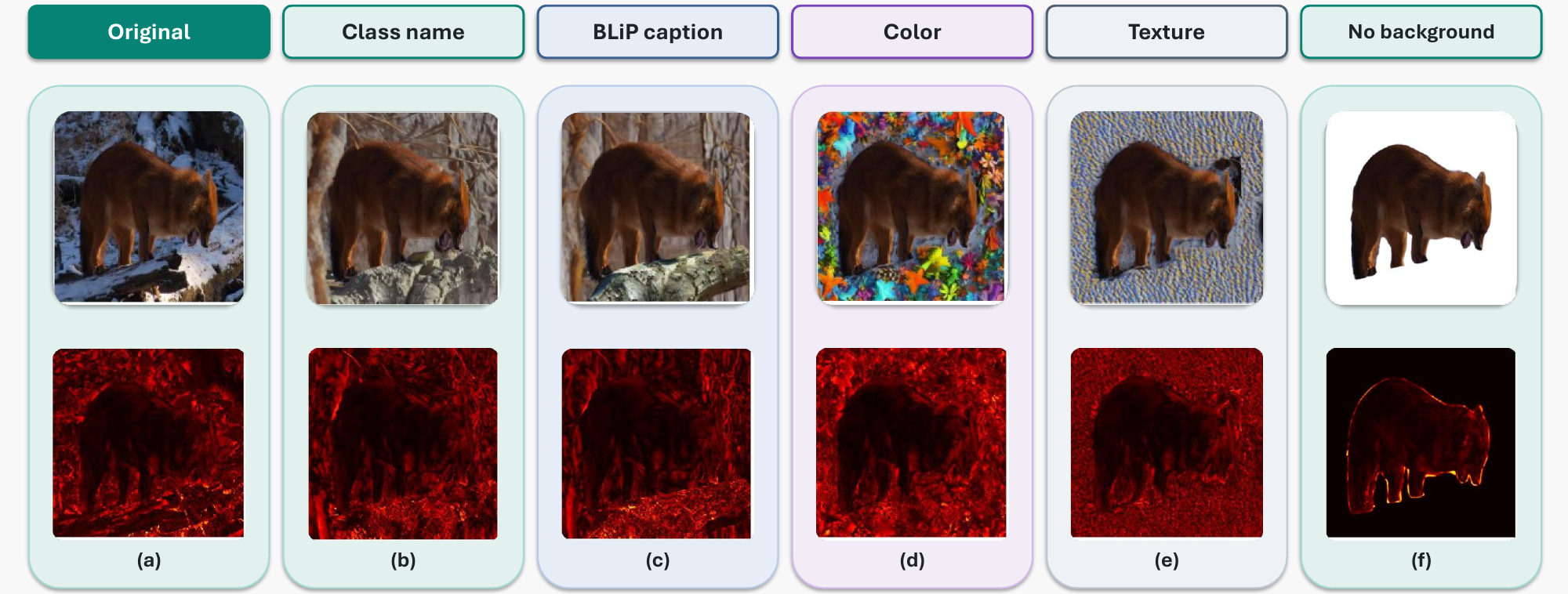}
	\caption{Reconstruction-error heatmaps across background conditions. Even in non-adversarial settings, error concentrates on the background. Under adversarial perturbation (color, texture), this background focus intensifies.}
	\label{fig:bg_error_heatmap}
\end{figure*}

\subsection{Background Bias Evaluation}

We report accuracy across six conditions as defined in Section~\ref{sec:preliminaries} (origin, class name, BLIP, color, texture, no-background). Table~\ref{tab:background} presents the results.

The diffusion classifier is substantially more sensitive to background perturbations in every condition. Under color and texture alterations, accuracy drops to 76.76\% and 72.38\% (a 15--19 point degradation from 91.43\%), while the baseline holds at 92.00\% and 90.10\%. The no-background condition is particularly revealing: removing the background recovers diffusion-classifier accuracy only to 81.30\%, still below its origin performance, whereas the baseline reaches 97.14\%, essentially matching its unaltered accuracy. This asymmetry indicates a structural entanglement between foreground recognition and surrounding context in the diffusion classifier.

\begin{table}[tb]
	\centering
	\caption{Background bias evaluation (top-1 accuracy, \%; higher is better).}
	\label{tab:background}
	\resizebox{\linewidth}{!}{%
	\begin{tabular}{lcccccc}
		\toprule
		Model & Origin & Class name & BLIP & Color & Texture & No BG \\
		\midrule
		Diffusion classifier & 91.43 & 96.95 & 84.95 & 76.76 & 72.38 & 81.30 \\
		OpenCLIP ViT-H/14   & 97.52 & 99.62 & 94.48 & 92.00 & 90.10 & 97.14 \\
		\bottomrule
	\end{tabular}}
\end{table}

\paragraph{Reconstruction-error heatmaps reveal a persistent background pull.}

Figure~\ref{fig:bg_error_heatmap} shows that even under non-adversarial conditions (origin, class name, BLIP), error already concentrates on the background rather than the foreground object, revealing background dominance as a baseline structural property. Under adversarial conditions (color, texture), this tendency intensifies further. Because prompt selection minimizes total pixel-aggregated error, background-weighted error distributions mean that large background regions dominate the classification signal regardless of foreground evidence.

\paragraph{Timestep analysis quantifies the object--background competition.}

We compute the object-to-background reconstruction-error ratio at each timestep (Appendix~\ref{appendix:bg_ratios}). For every condition with background present, the ratio stays below~1 throughout the entire denoising trajectory: background regions consistently absorb more error than the foreground object, and adversarial color/texture alterations intensify this imbalance further. For the no-background condition the ratio rises above~1 (error correctly concentrates on the object) but decreases monotonically at higher timesteps, indicating that background dependency grows with noise level. These analyses establish background dependency as a persistent structural property operative at every noise level and amplified under visual conflict.

\begin{takeaway}
On background dependency, the diffusion classifier relies \emph{far more} on context than the OpenCLIP baseline (15--19 point drops vs.\ ${\sim}2$), with the object-to-background error ratio below~1 at every timestep. This entanglement is intrinsic to reconstruction-error scoring, raising concerns about the reliability of diffusion classifiers.
\end{takeaway}

\subsection{\red{Generalization Across Architectures}}

{ To test whether these patterns are specific to the Stable Diffusion~2 U-Net, we replicate a representative subset of all three probes on Stable Diffusion~3, whose DiT backbone and flow-matching objective differ substantially; the same qualitative trends hold (Appendix~\ref{appendix:sd3}). Because SD2 and SD3 also differ in training data and scale, this evidence speaks to the \emph{generality} of these biases across diffusion families rather than isolating architecture as their cause, which we leave to future work.}


\section{Conclusion}
\label{sec:conclusion}

We introduced ASOB-Bench and conducted the first systematic assessment of how diffusion classifiers make decisions, probing their mechanisms across attribute binding, size-order, and background dependency dimensions against an OpenCLIP baseline sharing the same text encoder. Our results reveal a distinct, non-uniform decision-making profile: diffusion classifiers exhibit \emph{less} attribute binding bias across all tested categories, yet are substantially more susceptible to size-order shortcuts and background dependency. Reconstruction-error heatmaps and cross-attention visualizations provide mechanistic grounding for each finding. Because diffusion classifiers share their generative mechanism with text-to-image models, these insights suggest analogous failure modes in image generation, which our table-and-clock reconstruction illustrates qualitatively (Fig.~\ref{fig:full_recon}) and which we leave to systematic future study. Future work can exploit this knowledge to deploy diffusion classifiers more deliberately, steering them away from known shortcuts toward more robust and reliable predictions.

{
    \small
    \bibliographystyle{ieeenat_fullname}
    \bibliography{main}
}

\clearpage
\appendix

\section*{Appendix}

\noindent This appendix provides additional details and analyses that support the findings reported in the main paper. Specifically, it contains: (A) a systematic comparison of prompt formulations for two-object attribute binding evaluation and the rationale for the adopted prompt; (B) a timestep-resolved attention-ratio analysis for attribute binding bias, distinguishing structural cross-attention routing from co-occurrence-driven reconstruction errors; (C) a timestep-resolved error-ratio analysis for size-order bias, establishing object area---rather than word order---as the dominant driver of the Scenario~1 vs.\ Scenario~2 accuracy gap; (D) a timestep-resolved error-ratio analysis for background dependency bias, confirming background dominance as a persistent structural property across all denoising noise levels; \red{(E) a representative Stable Diffusion~3 (DiT) replication showing the same bias trends hold beyond the SD2 U-Net backbone; and (F) dataset statistics, the generation and quality-control pipeline, and definitions of all attribute categories}.

\section{Prompt Design for Two-Object Attribute Binding Evaluation}
\label{appendix:prompts}

The two-object prompt must acknowledge the distractor while expressing binding minimally. Table~\ref{tab:prompt_designs} presents all 12 candidate formulations with their CAB scores (reported on a $0$--$100$ scale; higher means more bias). We adopt \texttt{a \{color\} \{fruit\} and another object} (CAB\,$=$\,68.0) as it balances syntactic simplicity with explicit two-object structure. CAB spans more than 20 points across these formulations (60.5--82.0), so this choice is deliberately mid-range rather than the lowest-scoring one: the lower-CAB templates pad the prompt with scene-describing phrases (e.g., \emph{``in a complex scene''}) that dilute the two-object binding signal, whereas naming a single distractor minimally keeps the measurement focused on binding rather than on prompt verbosity. Reporting the full sweep also makes explicit that the reported bias is not an artifact of prompt selection. The CAB values in Table~\ref{tab:prompt_designs} are computed on a fixed subset of the natural-color data used only for prompt selection, so they differ slightly from the full-dataset numbers in the main paper (e.g., the adopted prompt scores 68.0 here versus 71.4 in the main paper); the relative ordering of the formulations is unaffected.

\begin{table}[tb]
\centering
\caption{All 12 two-object prompt formulations and their CAB scores ($\times 100$; higher $=$ more bias) on natural-fruit color binding. The adopted prompt is shown in \textbf{bold}.}
\label{tab:prompt_designs}
\small
\begin{tabular}{lc}
\toprule
Prompt Template & CAB \\
\midrule
a \{color\} \{fruit\} while another object appears nearby & 60.5 \\
a \{color\} \{fruit\} in a complex scene & 60.5 \\
a \{color\} \{fruit\} in a multi-object image & 61.0 \\
a \{color\} \{fruit\} among multiple objects & 61.5 \\
a \{color\} \{fruit\} & 62.5 \\
a \{color\} \{fruit\} with another item in the scene & 63.0 \\
a \{color\} \{fruit\} in an image with another object & 67.5 \\
\textbf{a \{color\} \{fruit\} and another object} & \textbf{68.0} \\
a \{color\} \{fruit\} alongside another object & 68.5 \\
a \{fruit\} with \{color\} coloring in a multi-object scene & 72.5 \\
in this picture, the color of the \{fruit\} is \{color\} & 78.0 \\
a \{fruit\} colored \{color\} alongside another object & 82.0 \\
\bottomrule
\end{tabular}
\end{table}

\section{Timestep Attention-Ratio Analysis for Attribute Binding}
\label{appendix:att_ratios}

\begin{figure*}[tb]
\centering
\begin{minipage}[t]{0.48\textwidth}
\centering
\includegraphics[width=\linewidth]{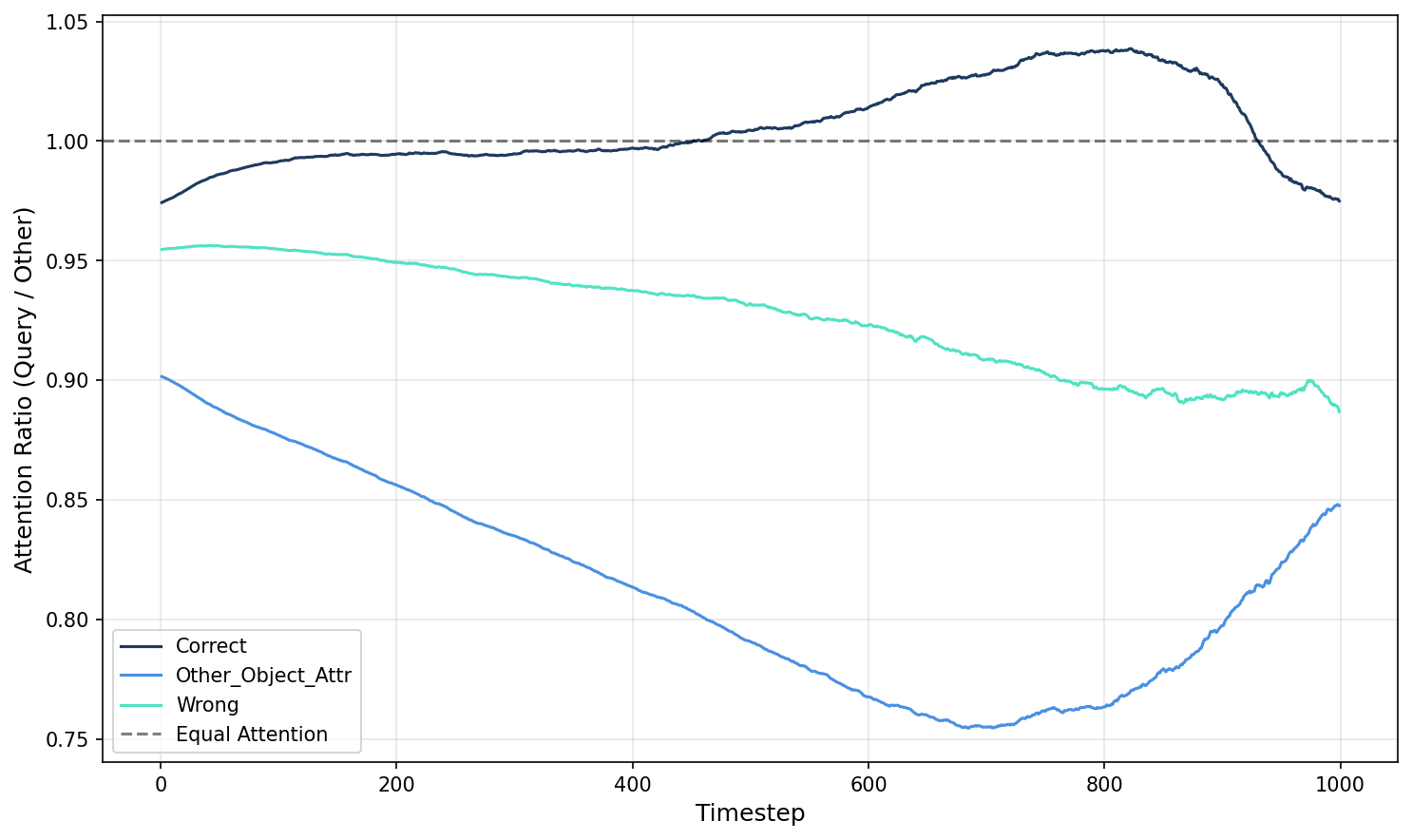}
\caption{Attention ratio (query/distractor) on natural images. The other-fruit-color prompt drives attention strongly toward the distractor (ratio well below 1).}
\label{fig:att_ratio_natural}
\end{minipage}\hfill
\begin{minipage}[t]{0.48\textwidth}
\centering
\includegraphics[width=\linewidth]{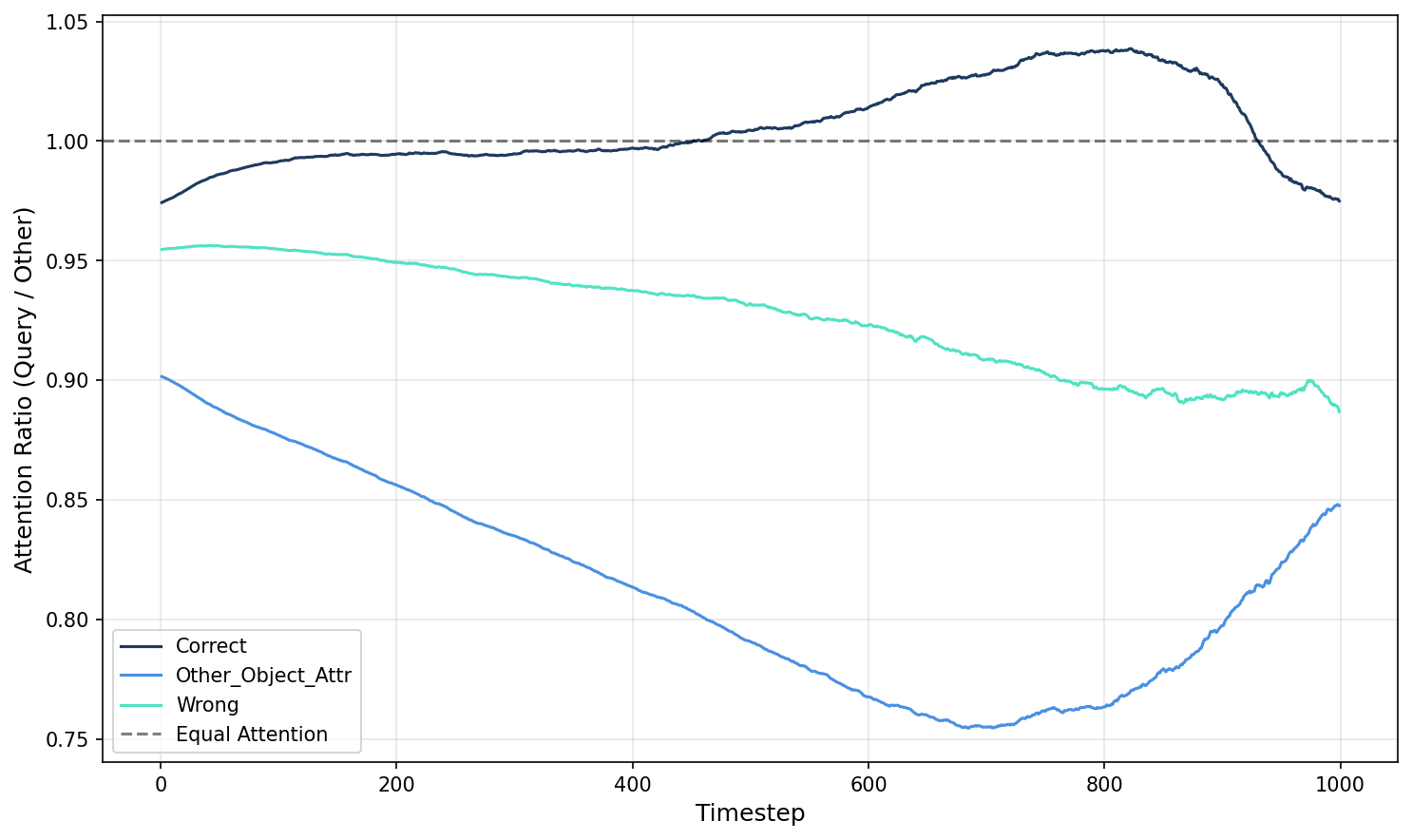}
\caption{Attention ratio on unnatural images. The pattern mirrors the natural case, confirming color-driven routing is intrinsic to the cross-attention mechanism.}
\label{fig:att_ratio_unnatural}
\end{minipage}
\end{figure*}

Figures~\ref{fig:att_ratio_natural} and~\ref{fig:att_ratio_unnatural} present the timestep-resolved attention-ratio analysis. For each timestep~$t$ we aggregate the U-Net cross-attention for the object token across all layers (following DAAM~\cite{Tang2023DAAM}) and compute the ratio of attention mass in the queried-object region to that in the distractor region; a ratio below~1 means attention is routed to the distractor rather than to the queried object. Cross-attention for the object token is driven primarily by the color cue in the prompt, and this pattern is nearly identical across natural and unnatural images, confirming that color-driven attention routing is an intrinsic property of the cross-attention mechanism independent of prototypical associations.

\section{Timestep Error-Ratio Analysis for Size-Order Bias}
\label{appendix:size_ratios}

For negative prompts, we compute for each timestep $t$ the ratio of total reconstruction error in the changed-object region to the average error in remaining objects' regions; a ratio above~1 indicates error correctly concentrates on the semantically changed region. For positive prompts, we compute the ratio of reconstruction error in the largest object's region to the average error in remaining regions, evaluated across both Scenario~1 and Scenario~2, to isolate the contribution of object area from word order.

\begin{figure*}[tb]
\centering
\begin{minipage}[t]{0.48\textwidth}
\centering
\includegraphics[width=\linewidth]{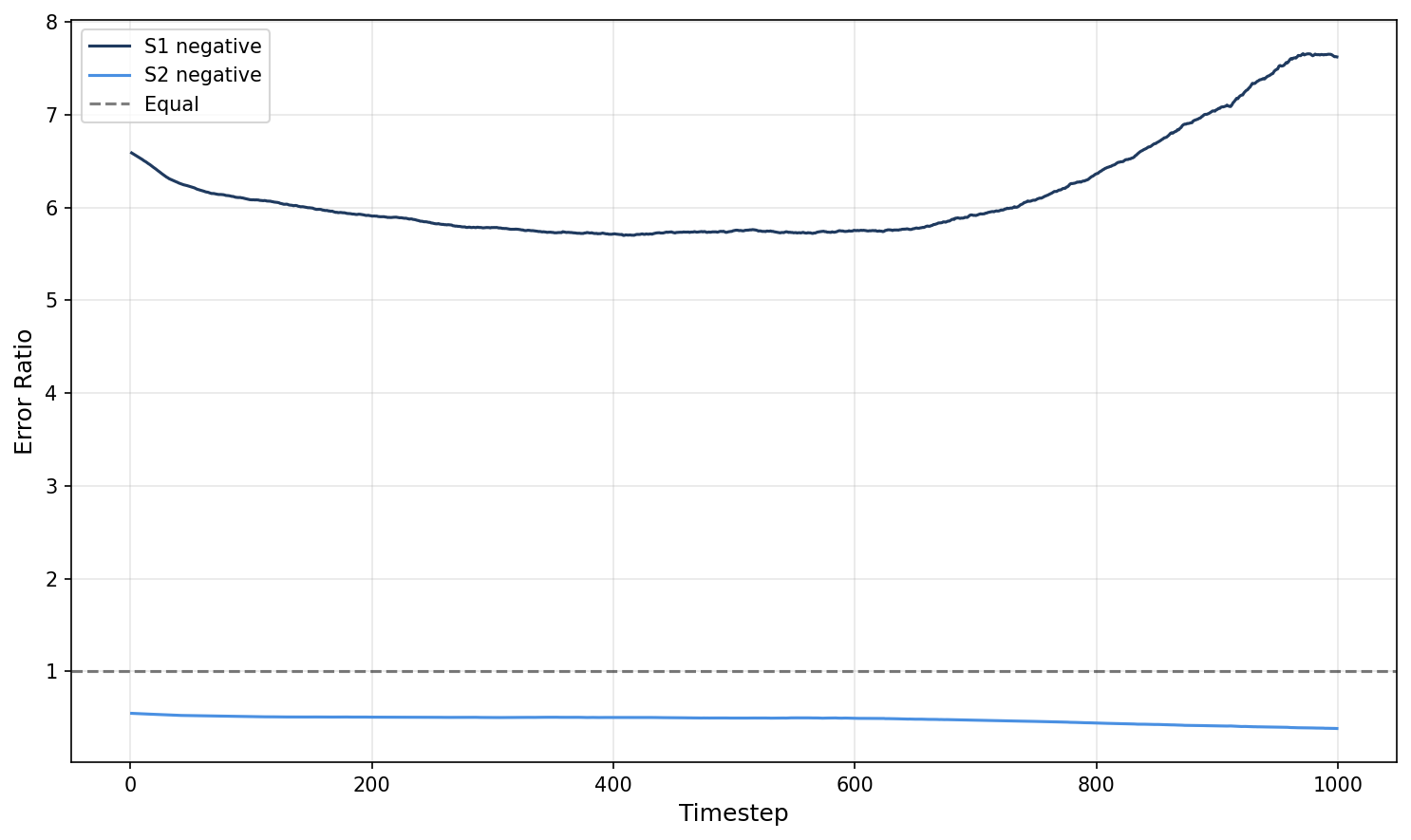}
\caption{Error ratio (changed object / remaining objects) for negative prompts. Scenario~1: ratio above~1. Scenario~2: ratio below~1, confirming the large unchanged object dominates the error budget.}
\label{fig:size_ratios_neg}
\end{minipage}\hfill
\begin{minipage}[t]{0.48\textwidth}
\centering
\includegraphics[width=\linewidth]{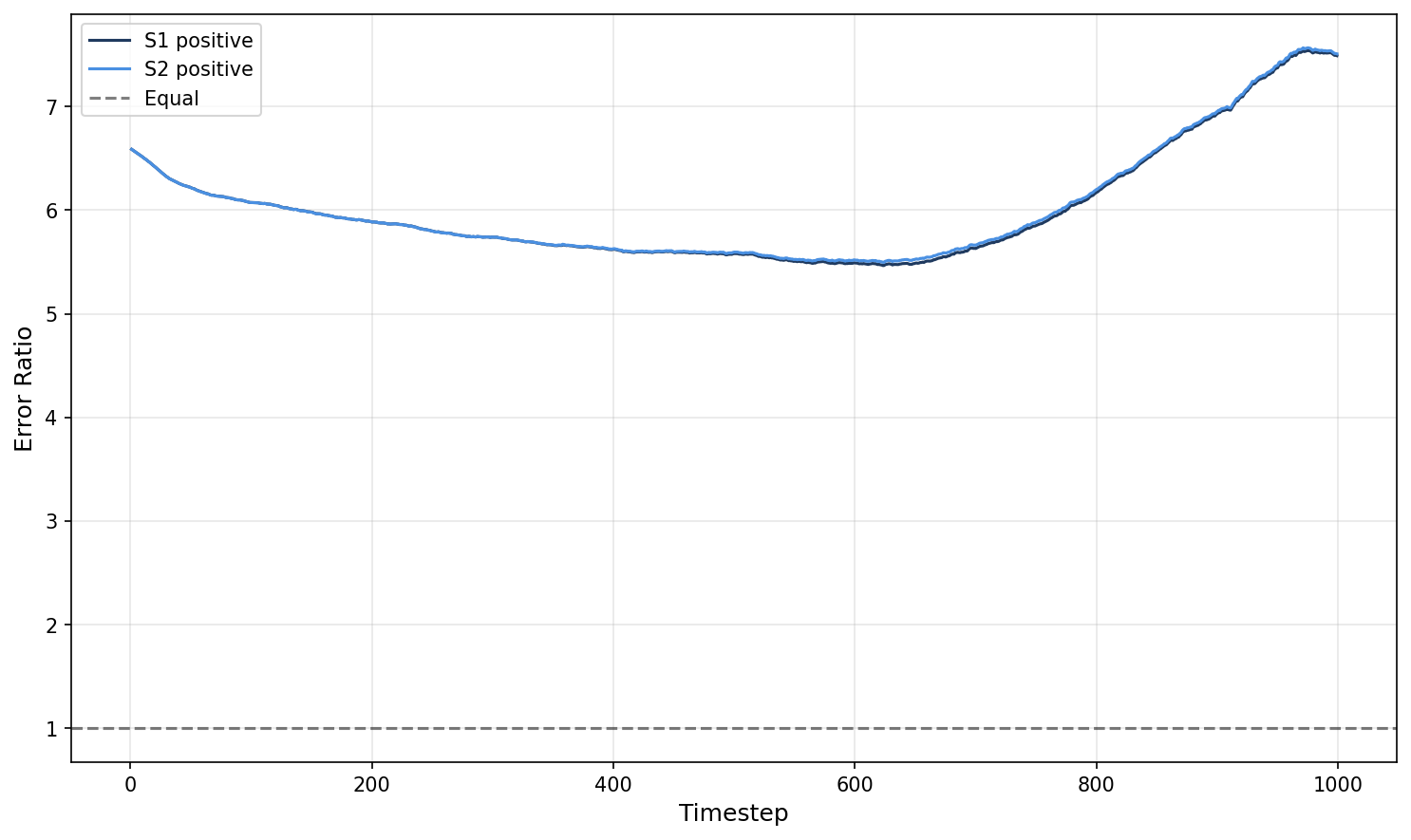}
\caption{Error ratio (biggest object / remaining objects) for positive prompts. Near-identical trajectories for Scenario~1 and Scenario~2 confirm word-order has negligible effect; object area is the dominant factor.}
\label{fig:size_ratios_pos}
\end{minipage}
\end{figure*}

\paragraph{Negative prompts (Figure~\ref{fig:size_ratios_neg}).}
When the changed object is the \emph{largest} (Scenario~1 negative), the ratio stays above~1, confirming the model correctly assigns elevated error to it. When the changed object is \emph{small} (Scenario~2 negative), the ratio drops below~1: the large unchanged object absorbs the dominant share of error by area alone, masking the semantically relevant change and causing classification accuracy to collapse.

\paragraph{Positive prompts (Figure~\ref{fig:size_ratios_pos}).}
The two curves follow nearly identical trajectories regardless of whether the largest object appears first or last in the prompt, demonstrating that error allocation is determined by image area rather than token position. Together, the two figures establish that the Scenario~1 vs.\ Scenario~2 accuracy gap is a structural consequence of size, not word order.

\section{Timestep Error-Ratio Analysis for Background Dependency}
\label{appendix:bg_ratios}

\begin{figure*}[tb]
\centering
\begin{minipage}[t]{0.48\textwidth}
\centering
\includegraphics[width=\linewidth]{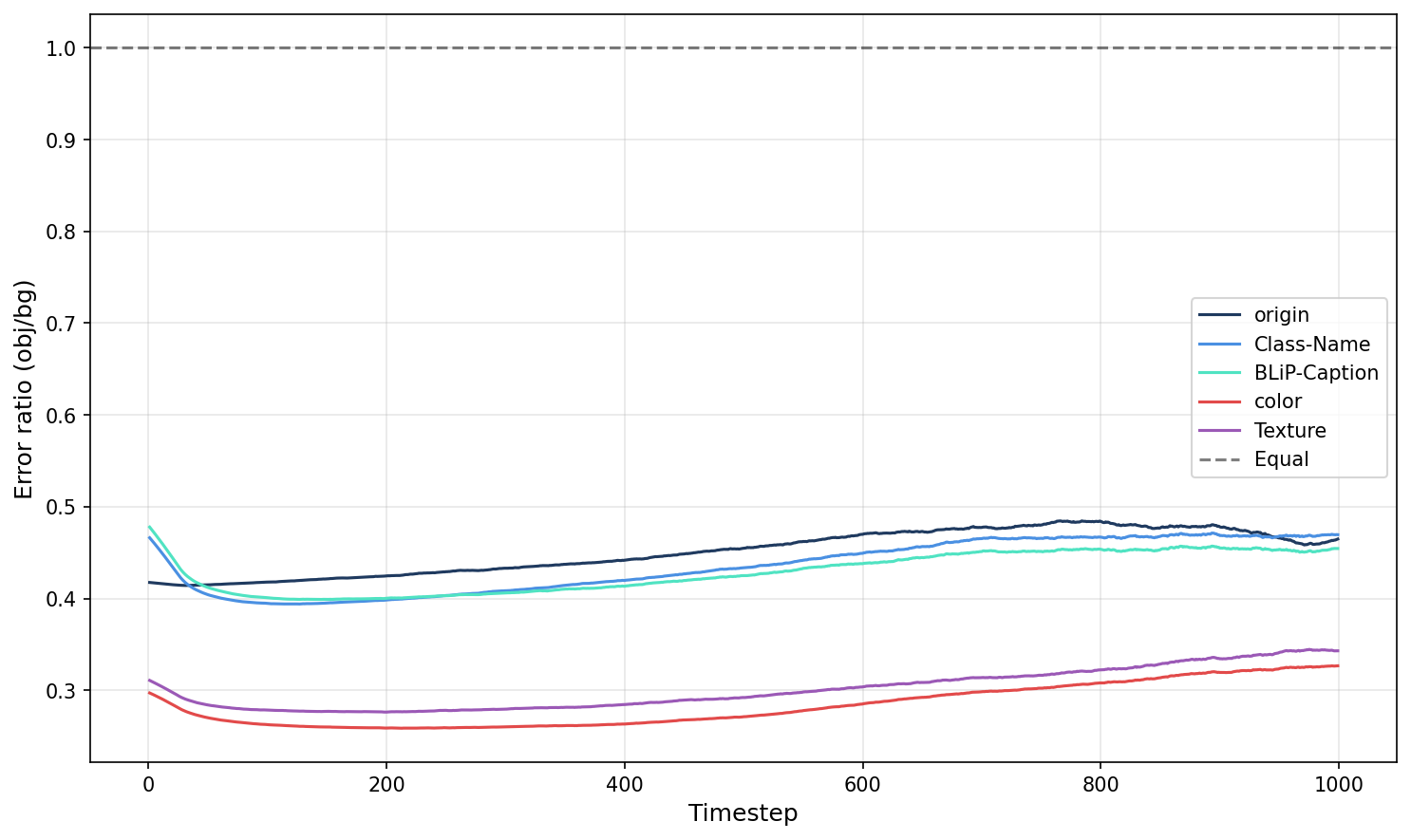}
\caption{Error ratio (object/background) for conditions with background. Ratio remains below 1 throughout: non-adversarial conditions cluster at 0.40--0.50; adversarial conditions (color, texture) fall to 0.25--0.35.}
\label{fig:bg_error_ratio}
\end{minipage}\hfill
\begin{minipage}[t]{0.48\textwidth}
\centering
\includegraphics[width=\linewidth]{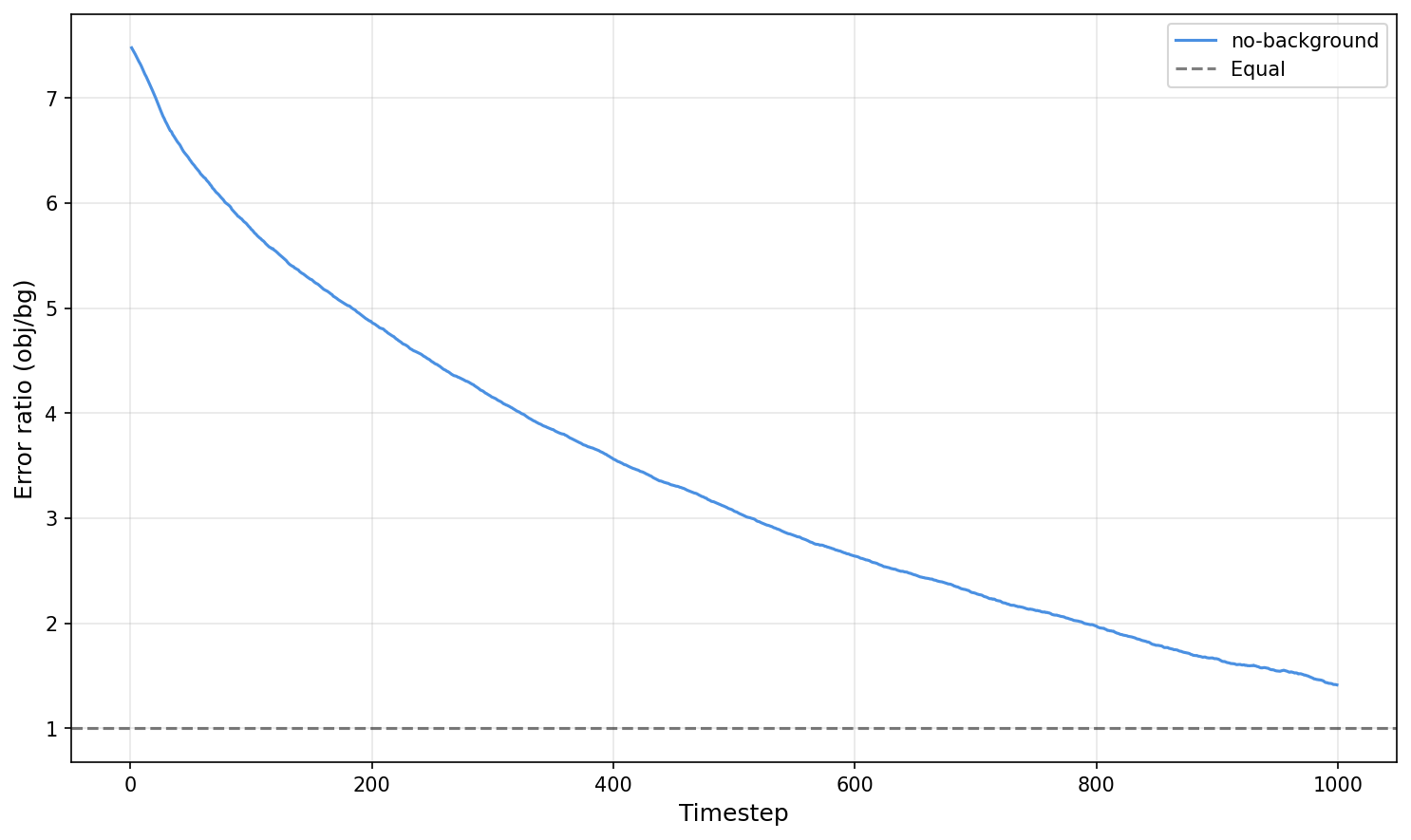}
\caption{Error ratio (object/background) for no-background condition. Ratio stays above 1 throughout but decreases monotonically at higher timesteps, reflecting increasing global noise structure.}
\label{fig:bg_error_ratio_nobg}
\end{minipage}
\end{figure*}

Figures~\ref{fig:bg_error_ratio} and~\ref{fig:bg_error_ratio_nobg} present the timestep-resolved error-ratio analysis for background dependency. The object-to-background ratio remains below 1 throughout the denoising process for all conditions with background present, confirming background dominance as a persistent structural property; when adversarial backgrounds are present, this ratio drops further, indicating that adversarial backgrounds intensify background dominance beyond the baseline. The no-background ratio stays above 1 but decreases monotonically at higher timesteps, indicating that background dependency increases as the denoising process progresses toward higher noise levels.

\section{\red{Generalization to Stable Diffusion 3 (DiT)}}
\label{appendix:sd3}

{
To verify that the biases reported in the main paper are not artifacts of the SD2 U-Net backbone, we replicate a representative subset of all three probes on Stable Diffusion~3 (SD3), which replaces the convolutional U-Net with a multimodal diffusion transformer (MM-DiT) trained under a flow-matching objective. Table~\ref{tab:sd3} reports the color-binding CAB, the two size-order scenarios (S1, S2), and background accuracy under the original and texture-corrupted conditions, alongside the corresponding SD2 numbers from the main paper for reference.

All three signatures persist: CAB remains well above the unbiased midpoint (attribute-binding bias), the Scenario~1$\rightarrow$Scenario~2 accuracy drop remains large (size-order shortcut), and background corruption sharply reduces accuracy (background dependency). The biases therefore generalize across two markedly different diffusion families.

We caution, however, that SD2 and SD3 differ not only in backbone architecture but also in training data, scale, and text-conditioning pipeline. The agreement in Table~\ref{tab:sd3} thus supports the \emph{generality} of these biases across diffusion models rather than isolating architecture as their cause, and absolute accuracies are not directly comparable across the two setups. We further note that SD3 is a weaker zero-shot classifier overall: its absolute accuracies are uniformly lower than SD2's (e.g., 42.10\% vs.\ 91.43\% on the original-background condition), consistent with prior reports that diffusion-classifier accuracy is highly sensitive to the backbone and timestep configuration~\cite{Jeong2025DiffusionCU}. The low SD3 numbers therefore reflect this general weakness rather than a malfunction; what matters here is that the \emph{relative} bias trends are preserved. A fully controlled ablation that varies only the backbone is left to future work.
}

\begin{table}[tb]
\centering
\caption{\red{Representative bias evaluation on SD3 (DiT) versus SD2 (U-Net). CAB is the color-binding diagnostic ($\times 100$; higher $=$ more bias); S1/S2 are size-order scenario accuracies (\%); Orig./Text.\ are background accuracy (\%) on the original and texture-corrupted conditions. The same qualitative trends hold for both models.}}
\label{tab:sd3}
{
\small
\begin{tabular}{lccccc}
\toprule
Model & CAB & S1 & S2 & Orig. & Text. \\
\midrule
SD2 (U-Net) & 71.40 & 95.28 & 58.53 & 91.43 & 72.38 \\
SD3 (DiT)   & 63.55 & 75.59 & 48.22 & 42.10 & 26.67 \\
\bottomrule
\end{tabular}
}
\end{table}

\section{\red{Dataset Statistics, Generation Pipeline, and Attribute Definitions}}
\label{appendix:datasets}

\begin{figure*}[tb]
\centering
\includegraphics[width=\textwidth]{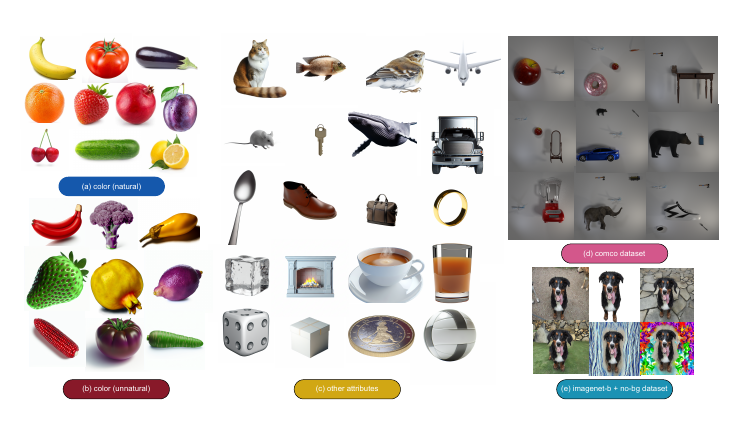}
\caption{Dataset samples. Panels (a)--(c) show \emph{single-object} images; the two-object scenes used in the color and attribute tasks are formed by pairing these programmatically (see the generation pipeline). (a)~natural-color fruits; (b)~unnatural-color DALL-E~3 fruits; (c)~additional attribute categories; (d)~ComCo multi-object scenes; (e)~ImageNet-B with original and altered backgrounds.}
\label{fig:datasets}
\end{figure*}

\begin{figure*}[tb]
\centering
\begin{minipage}[t]{0.19\textwidth}\centering
\includegraphics[height=1.55cm]{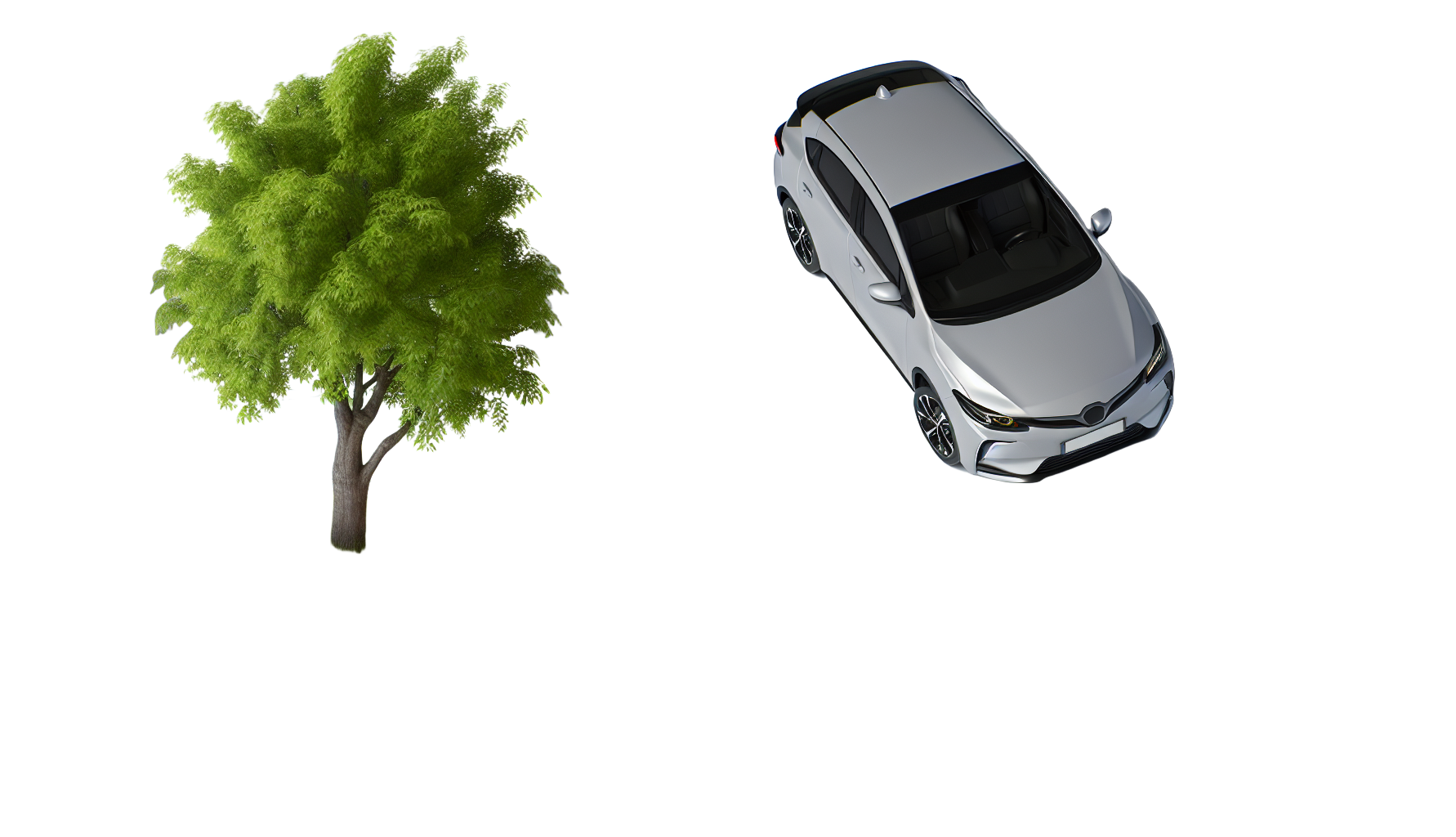}\\[2pt]
{\scriptsize (a)~Part--Whole}
\end{minipage}\hfill
\begin{minipage}[t]{0.19\textwidth}\centering
\includegraphics[height=1.55cm]{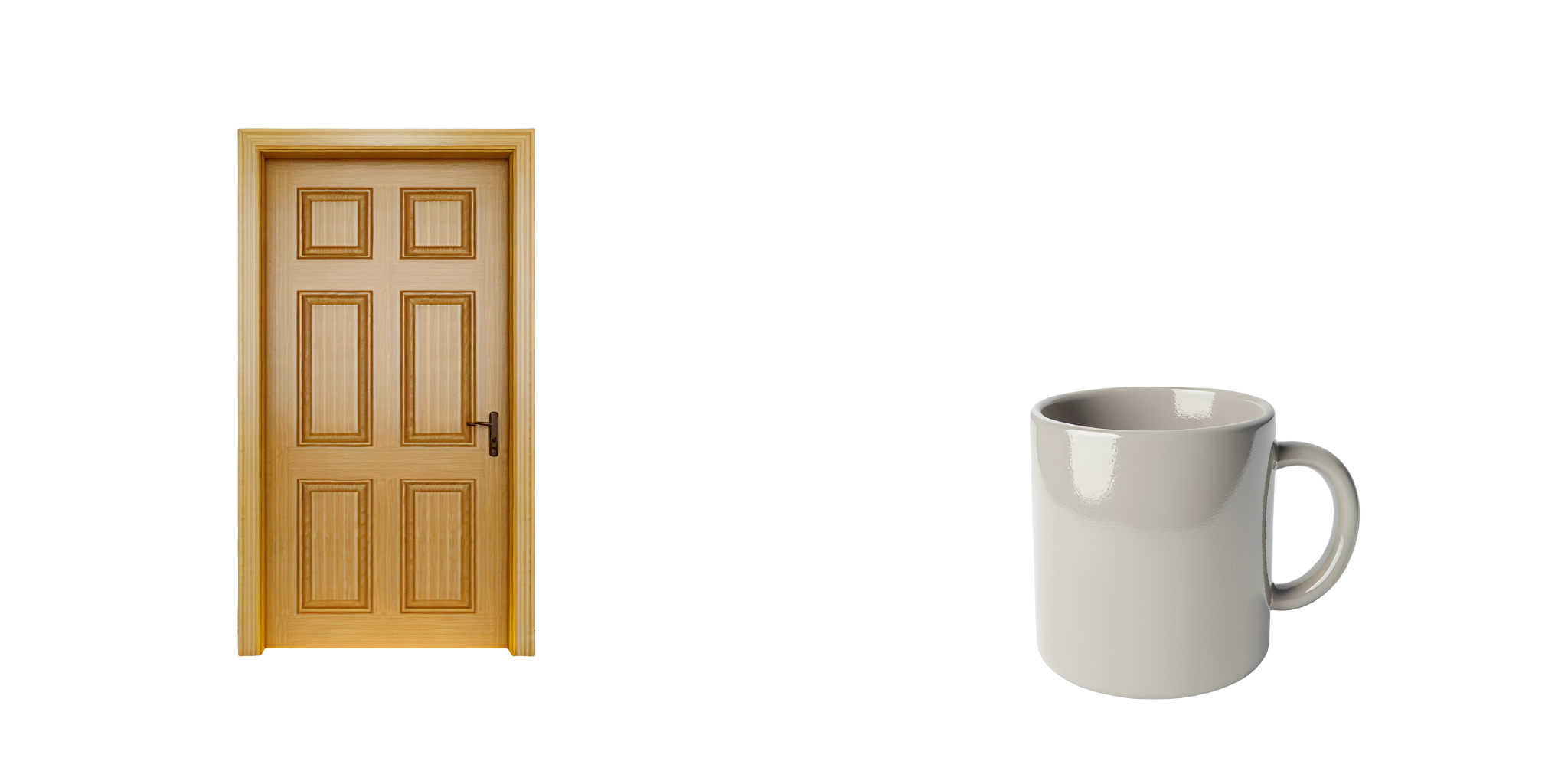}\\[2pt]
{\scriptsize (b)~Material/Texture}
\end{minipage}\hfill
\begin{minipage}[t]{0.19\textwidth}\centering
\includegraphics[height=1.55cm]{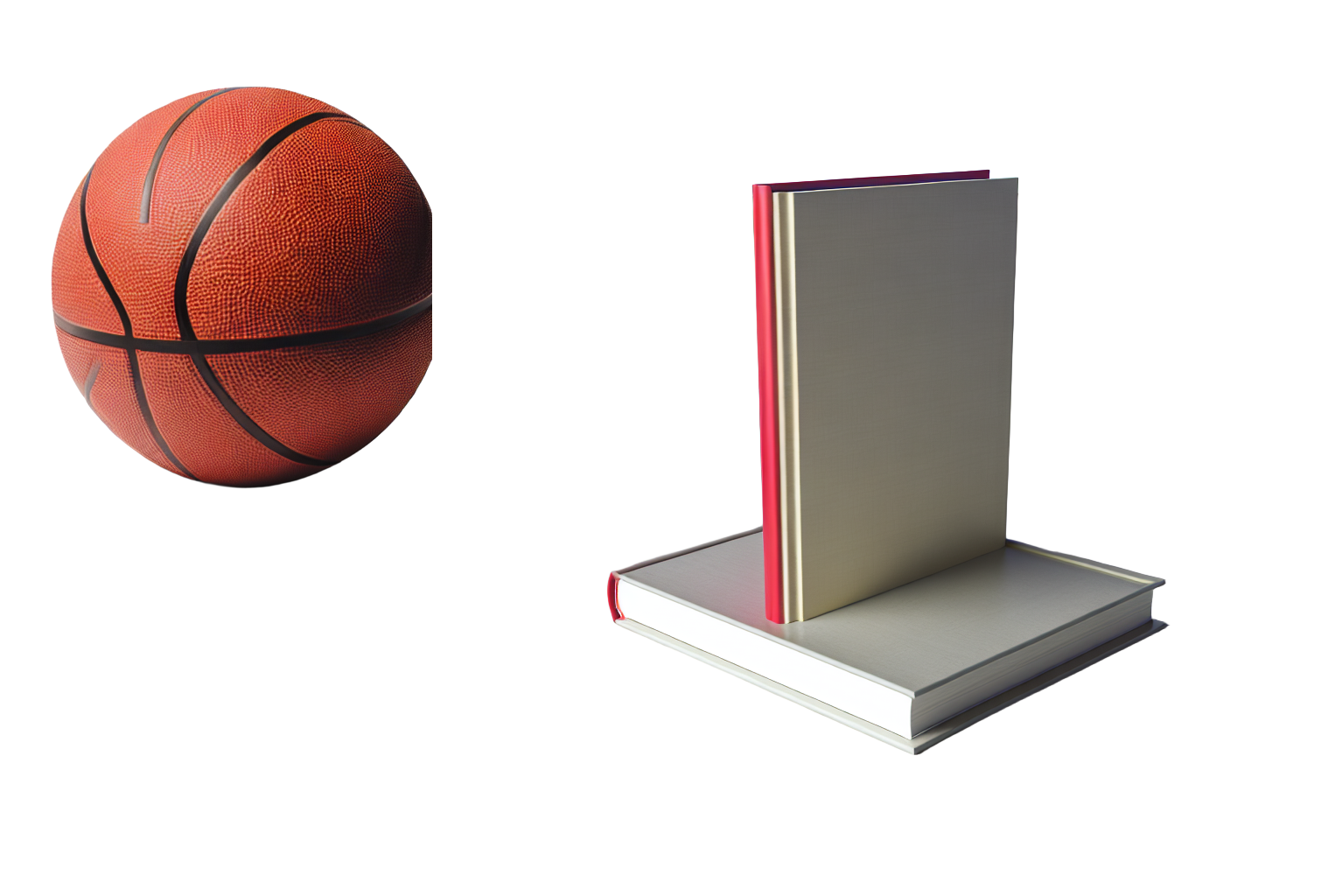}\\[2pt]
{\scriptsize (c)~Shape}
\end{minipage}\hfill
\begin{minipage}[t]{0.19\textwidth}\centering
\includegraphics[height=1.55cm]{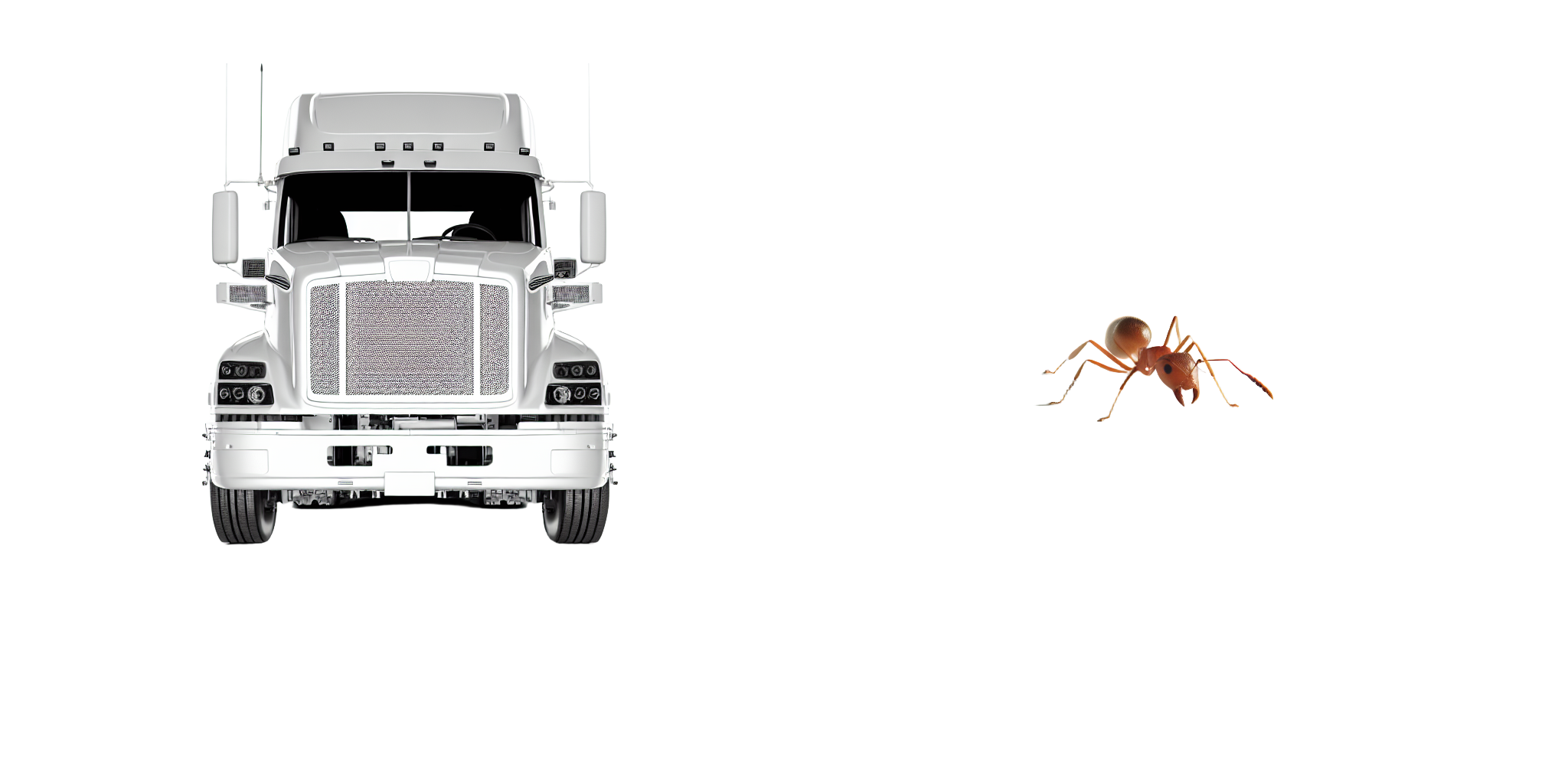}\\[2pt]
{\scriptsize (d)~Size}
\end{minipage}\hfill
\begin{minipage}[t]{0.19\textwidth}\centering
\includegraphics[height=1.55cm]{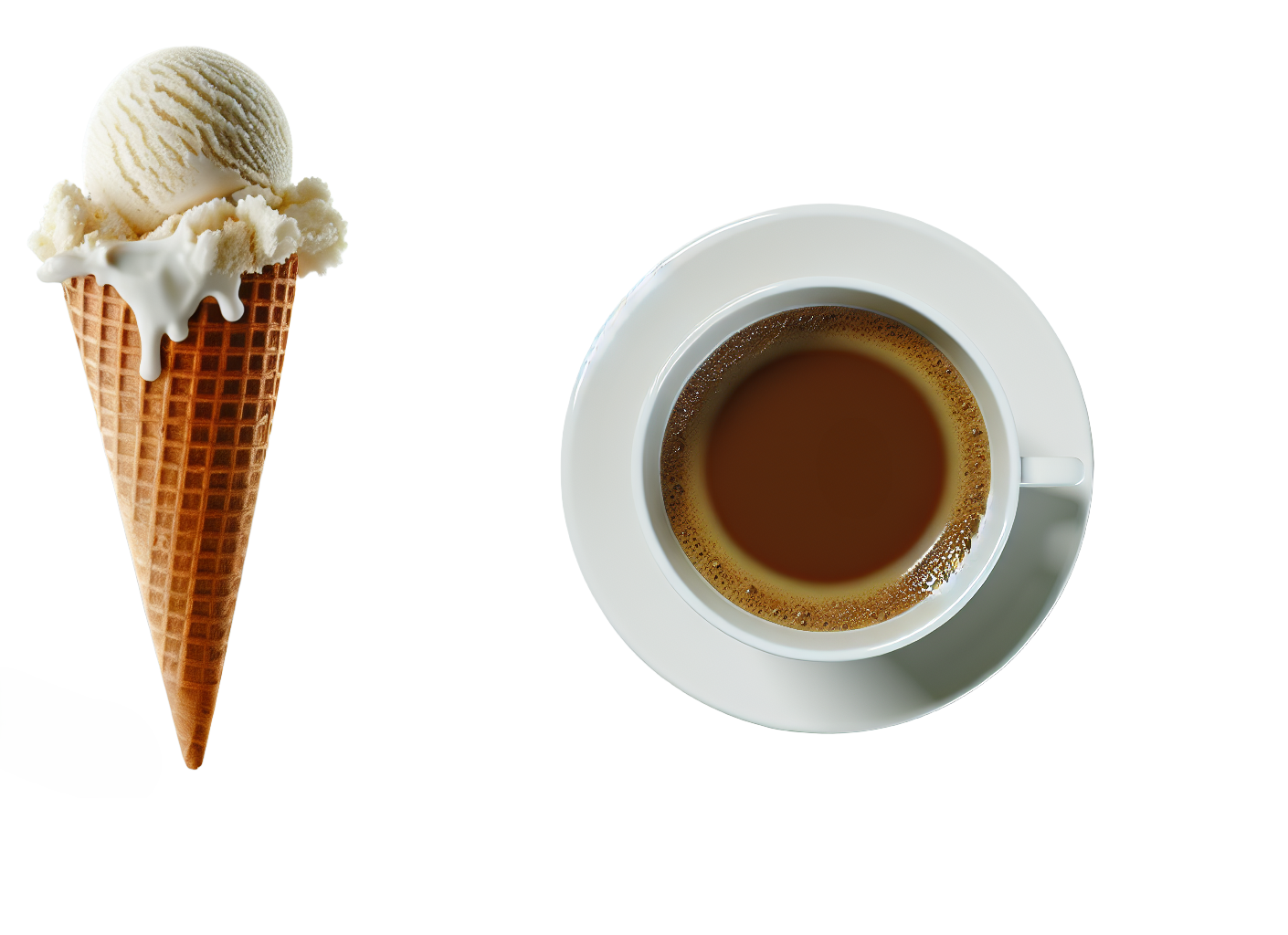}\\[2pt]
{\scriptsize (e)~Temperature}
\end{minipage}
\caption{\red{Representative two-object scenes for the five non-color attribute categories. Each scene pairs a queried object carrying the probed attribute with a distractor; the positive prompt takes the form ``a \{attribute\} \{object\} and another object'', and the corresponding negative prompt rebinds the distractor's attribute value to the queried object: (a)~a \emph{leafy} tree (vs.\ a \emph{wheeled} tree); (b)~a \emph{wooden} door (vs.\ a \emph{ceramic} door); (c)~a \emph{round} ball (vs.\ a \emph{square} ball); (d)~a \emph{big} truck (vs.\ a \emph{small} truck); (e)~a \emph{cold} ice cream (vs.\ a \emph{hot} ice cream).}}
\label{fig:attr_examples}
\end{figure*}

{
This appendix details the construction and composition of every dataset used in the main paper, separating the partitions adopted from prior work from those introduced here, and provides per-category sample counts, the generation and quality-control pipeline, and the definition of each attribute category. Figure~\ref{fig:datasets} shows representative samples from every dataset across the three bias dimensions.

\paragraph{Composition and provenance.} Table~\ref{tab:data_stats} lists each partition, its size, and its source. The natural-color fruit set is adopted from Tang et al.~\cite{Tang2022WhenAL}, and the ComCo size-order scenes~\cite{Abbasi2025CLIPUT} and the ImageNet-B background conditions~\cite{Malik2024ObjectComposeER} are adopted unchanged from prior work; the unnatural-color set and the five non-color attribute categories are constructed by us. The no-background variant is derived from ImageNet-B by segmenting foreground objects with BiRefNet~\cite{Zheng2024BiRefNet}.

\paragraph{Generation pipeline.} All synthetic probes are built from \emph{single-object} images generated with DALL-E~3 (Figure~\ref{fig:datasets}, (a)--(c)) and then composed into two-object scenes programmatically. The unnatural-color fruits are generated with the template
\begin{quote}\small\ttfamily
A highly realistic photograph of a single \{color\} \{fruit\}. The fruit should look natural and fresh, with realistic texture, captured in professional food photography style, with natural lighting.
\end{quote}
where the (unnatural) color is specified explicitly. The five additional attribute categories use
\begin{quote}\small\ttfamily
A realistic, high-quality image of a \{object\} centered on a plain white background, studio lighting, no shadows, no text, no people, minimalistic composition.
\end{quote}
Here the attribute is \emph{not} named in the prompt: because we probe each object's \emph{prototypical} attribute, the generator renders the intended attribute in the large majority of cases (e.g., tables are almost always square), so specifying the object alone suffices.

\paragraph{Quality control and pairing.} Each generated single-object image was manually screened and \emph{discarded} (not regenerated) if it was low quality or did not exhibit the intended attribute. Surviving images were then paired by code into two-object scenes such that the two objects \emph{and} their attribute values both differ (e.g., a \{color\,1\} \{fruit\,1\} placed beside a \{color\,2\} \{fruit\,2\}, or two objects carrying different shapes), producing the queried-object/distractor structure used throughout. Our natural-color set is real, adopted from Tang et al.~\cite{Tang2022WhenAL}. Only the unnatural-color set and the five non-color attribute categories are synthesized with DALL-E~3. The reported bias is therefore not specific to generated data, since it already appears clearly on the real natural-color set. We nonetheless verify that the biases measured on the synthetic probes are not artifacts of generation.

Generation can corrupt the measurement in two concrete ways, and we control both. First, the generator might not depict the intended attribute, which would corrupt the label. We therefore screen every single-object image manually and discard any that lacks the attribute. Second, the generator might add layout cues, such as making one object larger or more salient. We avoid this by composing the two-object scenes with code, so the spatial arrangement is set by us, not the generator.

The remaining worry is that generated images carry a hidden visual signature tied to the attribute, which the classifier could exploit. The bias itself rules this out. In every category the bias is the same effect: the classifier defaults to each object's \emph{prototypical} attribute. The unnatural-color condition tests this directly. These images come from the same pipeline, but the prototypical color--object link is broken, and the CAB gap then disappears (Table~\ref{tab:attribute_results}). If such a signature were driving the bias, it would still be present in the unnatural-color images, because they come from the same pipeline, and the bias should remain. Instead, the bias vanishes. The only thing that changed is the meaning of the image, since the color is no longer prototypical for the object. The cause is therefore a learned semantic prior, not a hidden generation signature. Because the same prototypical-default effect underlies every attribute category, this conclusion covers all of them.

\paragraph{Attribute category definitions.} Table~\ref{tab:attr_defs} defines each attribute category and gives a representative example, and Figure~\ref{fig:attr_examples} shows a representative two-object scene for each of the five non-color categories. Every category follows the same two-object structure as the color task, pairing a queried object carrying the attribute with a distractor.
}

\begin{table}[tb]
\centering
\caption{\red{Dataset partitions, sizes, and provenance. ``\# samples'' is the number of samples actually \emph{used} in our evaluation, not the full size of each source dataset (e.g., for ComCo and ImageNet-B we evaluate on a subsample, since scoring one image requires a noise-prediction pass for every candidate prompt over 50 timesteps). Synthetic partitions (unnatural color and the five attribute categories) are generated with DALL-E~3; ImageNet-B supplies five background conditions. ``Ours'' denotes data introduced in this work. All partitions are balanced-sampled across their categories, except ImageNet-B, which retains only the 20 classes with the highest image quality.}}
\label{tab:data_stats}
{
\small
\begin{tabular*}{\linewidth}{@{\extracolsep{\fill}}lcc@{}}
\toprule
Partition & \# samples used & Source \\
\midrule
Color (natural)   & 1{,}500 & Adopted~\cite{Tang2022WhenAL} \\
Color (unnatural) & 669    & Ours (synthetic) \\
Part--Whole       & 600    & Ours (synthetic) \\
Material/Texture  & 600    & Ours (synthetic) \\
Shape             & 600    & Ours (synthetic) \\
Size              & 600    & Ours (synthetic) \\
Temperature       & 600    & Ours (synthetic) \\
ComCo             & 12{,}780 & Adopted~\cite{Abbasi2025CLIPUT} \\
ImageNet-B        & 2{,}625 & Adopted~\cite{Malik2024ObjectComposeER} \\
No-background     & 525    & Derived~\cite{Malik2024ObjectComposeER,Zheng2024BiRefNet} \\
\bottomrule
\end{tabular*}
}
\end{table}

\begin{table}[tb]
\centering
\caption{\red{Definitions of the attribute categories with representative two-object examples.}}
\label{tab:attr_defs}
{
\small
\begin{tabular}{lp{0.56\linewidth}}
\toprule
Category & Attribute probed and example \\
\midrule
Color            & Surface color; e.g., ``an orange carrot and another object''. \\
Part--Whole      & Whether a characteristic part binds to the queried object; e.g., ``a leafy tree and another object''. \\
Material/Texture & Material/surface; e.g., ``a wooden chair and another object''. \\
Shape            & Geometric shape; e.g., ``a square table and another object''. \\
Size             & Size adjective; e.g., ``a big elephant and another object''. \\
Temperature      & Thermal attribute; e.g., ``a hot coffee and another object''. \\
\bottomrule
\end{tabular}
}
\end{table}

\end{document}